\documentclass[11pt, letterpaper, logo, twocolumn]{noitom_robotics}

\usepackage{amsmath}
\usepackage{amssymb}
\usepackage{times}
\usepackage{dsfont}
\usepackage{xspace}
\usepackage{wrapfig}
\usepackage{float}
\usepackage{stfloats}
\usepackage{capt-of}
\usepackage{subfigure}
\usepackage{listings}
\usepackage{multirow}
\usepackage{makecell}
\usepackage[flushleft]{threeparttable}
\usepackage{arydshln}
\usepackage[table,xcdraw]{xcolor}

\definecolor{ourcolor}{HTML}{EE6363}
\definecolor{darkred}{HTML}{B35446}
\hypersetup{allcolors=ourcolor}

\newcommand{\cmark}{\textcolor{green}{\ding{51}}}
\newcommand{\xmark}{\textcolor{red}{\ding{55}}}
\newcommand{\ours}{\textcolor{ourcolor}{OmniContact}\xspace}

\title{\ours: Chaining Meta-Skills via Contact Flow for Generalizable Humanoid Loco-Manipulation}
\newcommand{\website}{https://omnicontact.github.io/}
\runningtitle{\ours: Chaining Meta-Skills via Contact Flow}
\renewcommand{\today}{June 2026}

\author[1,2,*]{Runyi Yu}
\author[1,3,*]{Xiaoyi Lin}
\author[1]{Ji Ma}
\author[2,\ding{41}]{Yinhuai Wang}
\author[2]{Koukou Luo}
\author[1]{Jiahao Ji\protect\\}
\author[1,4]{Huayi Wang}
\author[1,4]{Wenjia Wang}
\author[1]{Runhan Zhang}
\author[2]{Ping Tan}
\author[1]{Ting Wu\protect\\}
\author[1]{Ruoli Dai}
\author[2,\ding{41}]{Qifeng Chen}
\author[1,\ding{41}]{Lei Han}

\affil[1]{Noitom Robotics,\quad}
\affil[2]{The Hong Kong University of Science and Technology\protect\\}
\affil[3]{Wuhan University,\quad}
\affil[4]{The University of Hong Kong}

\manualauthors{%
Runyi Yu\textsuperscript{1,2,*},\quad
Xiaoyi Lin\textsuperscript{1,3,*},\quad
Ji Ma\textsuperscript{1},\quad
Yinhuai Wang\textsuperscript{2,\ding{41}},\quad
Koukou Luo\textsuperscript{2},\quad
Jiahao Ji\textsuperscript{1},\\
Huayi Wang\textsuperscript{1,4},\quad
Wenjia Wang\textsuperscript{1,4},\quad
Runhan Zhang\textsuperscript{1},\quad
Ping Tan\textsuperscript{2},\quad
Ting Wu\textsuperscript{1},\\
Ruoli Dai\textsuperscript{1},\quad
Qifeng Chen\textsuperscript{2,\ding{41}},\quad
Lei Han\textsuperscript{1,\ding{41}}\\[0.6em]
\Affilfont
\textsuperscript{1}Noitom Robotics,\quad
\textsuperscript{2}The Hong Kong University of Science and Technology\\
\textsuperscript{3}Wuhan University,\quad
\textsuperscript{4}The University of Hong Kong%
}

\correspondingauthor{Yinhuai Wang, Qifeng Chen, Lei Han}
\paperurl{https://omnicontact.github.io/}
\keywords{Humanoid Loco-Manipulation, Long-Horizon Execution}

\begin{document}
\maketitle

\section*{Abstract}
Learning long-horizon humanoid loco-manipulation poses a dual challenge: it requires not only the robust execution of meta-skills but also their seamless, closed-loop chaining equipped with autonomous recovery.
Existing approaches remain limited: explicit humanoid-object interaction representations offer precision but are notoriously difficult for high-level planning, whereas implicit skill embeddings are compact but lack the interpretability required for reliable composition.
We propose \ours, a hierarchical framework centered on \textbf{contact flow (CF)}, a compact representation consisting of key body trajectories and time-series binary contact signals. Leveraging this shared interface, our low-level policy \textbf{CF-Track} learns a unified library of loco-manipulation skills, while our high-level module \textbf{CF-Gen} heuristically synthesizes future contact-flow sequences. To support this setting, we additionally collect the OmniContact dataset, a MoCap-based HOI corpus for humanoid loco-manipulation (Appendix~\ref{sec:dataset}). Together, they enable robust execution, autonomous failure recovery, and flexible composition of meta-skills for long-horizon tasks. 
Experiments show that OmniContact achieves \(98.7\%\) success on \textit{Carry Box} and \(76.5\%\) on \textit{Push-Stack Boxes}, outperforming prior baselines by average margins of \(40.9\%\) in meta-skill and \(66.5\%\) in skill chaining. Besides, our framework naturally integrates with VLMs for semantic task decomposition, enabling complex, semantically grounded loco-manipulation behaviors, such as arranging scattered boxes into a heart shape.

\section{Introduction}
\label{sec:intro}

\begin{table*}[t]
    \centering
    \small
    \captionsetup{skip=2pt}
    \caption{\textbf{Comparison with representative methods}.
    Here, TT denotes tracking target. Object-pose generalization refers to the ability to generalize across different initial and target object poses.}
    \label{tab:comparison}
    \setlength{\tabcolsep}{4pt}
    \renewcommand{\arraystretch}{1.2}
	\resizebox{\linewidth}{!}{
    \begin{tabular}{l | c c c c | c c c c}
        \toprule
        \multirow{2}{*}{Method} & \multicolumn{4}{c|}{Model Properties} & \multicolumn{4}{c}{Model Capabilities} \\
        \cmidrule(lr){2-5} \cmidrule(lr){6-9}
        & Policy & Task & Skill Rep. & Obj. Perception & Unified & Obj.-Pose Gen. & Skill Chaining & Recovery \\
        \midrule
        SONIC~\citep{luo2025sonic} & RL & Body Motion & Body TT & -- & \cmark & \xmark & \xmark & \xmark \\
        HumanPlus~\citep{fu2024humanplus} & BC & Interaction & -- & Vision & \cmark & \xmark & \xmark & \xmark \\
        VIRAL~\citep{he2025viral} & BC & Interaction & -- & Vision & \cmark & \xmark & \xmark & \xmark \\
        HDMI~\citep{weng2025hdmi} & RL & Interaction & HOI TT & 6D Pose & \cmark & \xmark & \xmark & \xmark \\
        Omniretarget~\citep{yang2025omniretarget} & RL & Interaction & HOI TT & 6D Pose & \cmark & \xmark & \xmark & \xmark \\

        HumanX~\citep{wang2026humanx} & RL & Interaction & - & 6D Pose & \cmark & \cmark & \xmark & \cmark \\
        PhysHSI~\citep{wang2025physhsi} & RL & Interaction & Object Goal & 6D Pose & \xmark & \cmark & \xmark & \cmark \\
        LessMimic~\citep{lin2026lessmimic} & RL & Interaction & Skill Embed. & 6D Pose & \xmark & \cmark & \cmark & \cmark \\
        \midrule
        \textbf{OmniContact} & RL & Interaction & Contact Flow & 6D Pose & \cmark  & \cmark & \cmark & \cmark \\
        \bottomrule
    \end{tabular}
}
\end{table*}

Enabling humanoid robots to autonomously solve robust, long-horizon loco-manipulation tasks remains a fundamental challenge in robotics, as it requires seamlessly coordinating whole-body motion with continuous object interaction, such as carrying boxes, pushing suitcases, or kicking objects~\cite{shi2026egohumanoid, wang2025physhsi, li2026haic, weng2025hdmi, yin2025visualmimic, lin2026lessmimic, wu2026sugar, liu2025opt2skill, he2025viral, xue2025opening, dong2026learning, he2024omnih2o, zhang2025wococo, sun2025ulc}. 
Beyond merely executing isolated skills, the fundamental challenge lies in composing these primitives into robust, closed-loop behaviors that can adapt to dynamic environments and recover autonomously from failures.

As summarized in Table~\ref{tab:comparison}, despite remarkable progress in humanoid loco-manipulation, existing approaches still fall short of this goal. 
Behavior-cloning-based methods~\citep{fu2024humanplus, shi2026egohumanoid} can acquire complex skills from human demonstrations, but their reliance on expert operators makes them difficult to scale and often leads to slow, open-loop execution. 
Alternatively, reinforcement learning (RL) offers a promising route to robust control, but existing RL-based methods remain limited. We broadly group them into three paradigms: (1) body motion learning, (2) HOI tracking, and (3) task-specific HOI learning. Body motion learning methods~\cite{ben2025homie,luo2025sonic,ze2025twist} often rely on teleoperated body motion and manipulate objects in an open-loop manner, making recovery from interaction failures difficult. HOI tracking approaches~\citep{weng2025hdmi, yang2025omniretarget} incorporate object perception, but depend on dense frame-by-frame whole-body references, which limits generalization and autonomous recovery. Task-specific HOI learning methods~\citep{wang2025physhsi,wang2026humanx} show strong robustness and recovery, but they are typically restricted to isolated skills and therefore provide limited support for skill composition in long-horizon tasks.

Taken together, these limitations reveal a deeper challenge: chaining meta-skills for composition and closed-loop execution in long-horizon tasks. Realizing such flexible composition requires answering a fundamental question: \textbf{\textit{what is the optimal representation for skill reuse?}} 
Existing choices fall short: full human-object interaction (HOI) states~\cite{wang2023physhoi,weng2025hdmi,yang2025omniretarget} are precise but complex for planning, whereas implicit skill embeddings~\citep{lin2026lessmimic, yu2025skillmimic, wang2025skillmimic} are compact but difficult to interpret, making structured skill composition challenging.
To resolve this dilemma, we look to the physical essence of the tasks.
We posit that the fundamental distinction between loco-manipulation and locomotion lies in object contact dynamics. Based on this insight, we propose \textbf{contact flow (CF)}, a compact and expressive representation consisting of key body trajectories and a time-series binary contact signal. Contact flow is expressive enough to capture diverse manipulation intents (e.g., carrying, pushing, and kicking), yet structured enough to facilitate heuristic synthesis and efficient high-level planning.

Centered around contact flow, we propose \ours, a complete system for closed-loop long-horizon humanoid loco-manipulation. This system consists of two modules bridged by this shared representation: (1) \textbf{CF-Track} serves as a low-level controller that learns a diverse library of interaction skills under a unified imitation-learning framework, with contact flow as the common skill input. (2) \textbf{CF-Gen} acts as a mid-level planner that synthesizes future contact-flow sequences from object-centric rules and contact anchors. To support this unified learning setup, we additionally collect the OmniContact dataset, a MoCap-based HOI corpus spanning diverse loco-manipulation primitives (Appendix~\ref{sec:dataset}).
With CF-Track and CF-Gen, OmniContact enables seamless chaining of meta-skills for the robust execution of long-horizon tasks and autonomous recovery.

Extensive experiments demonstrate that our method significantly outperforms prior baselines across three levels of complexity: (1) individual meta-skills, such as box carrying; (2) long-horizon tasks, such as box stacking; and (3) skill compositions, such as seamlessly combining carrying and pushing behaviors. 
Furthermore, in a stress test of its extreme long-horizon endurance, we found that \ours can execute continuous box-carrying tasks for around 40 minutes (Appendix~\ref{sec:append_long_horizon}).
Besides, our method also exhibits strong recovery abilities. Facilitated by real-time heuristic synthesis, CF-Gen can rapidly detect failures and replan accordingly, enabling the robot to swiftly re-approach an accidentally dropped object and seamlessly resume the task.
In addition, OmniContact is naturally compatible with high-level vision-language reasoning. By prompting VLMs to decompose complex tasks and output start-to-goal object poses, CF-Gen can automatically synthesize continuous contact flows using contact anchor templates. This integration empowers the robot to solve complex, semantically grounded tasks, such as arranging scattered boxes into a heart shape or sorting diverse objects into designated semantic bins. 
Ultimately, these results highlight our hierarchical framework as an effective bridge connecting high-level semantic planning, mid-level heuristic synthesis, and low-level skill execution. 

We summarize our contributions as follows:
\begin{itemize}
    \item \textbf{A Contact-Centric Skill Representation:} We propose \textbf{Contact Flow}, a compact and expressive representation combining body keypoint trajectories and binary contact signals, to capture the core dynamics of loco-manipulation for skill reuse and planning.
    \item \textbf{A Unified Closed-Loop System:} We develop \textbf{OmniContact}, a hierarchical framework that integrates CF-Track for robust skill execution and CF-Gen for heuristic contact-flow synthesis, enabling seamless chaining of meta-skills for long-horizon tasks. Our method outperforms baselines in meta-skill learning, enabling extreme long-horizon endurance ($\sim$40 minutes), autonomous recovery, and seamless VLM integration for complex tasks.
    \item \textbf{Dataset Release and Scalability:} We contribute a diverse MoCap-based HOI dataset (Appendix~\ref{sec:dataset}) and demonstrate positive scaling of robustness and generalization with data size.
\end{itemize}

\section{Related Work}
\label{sec:related_work}

\subsection{Reinforcement Learning for Humanoid Skills}
Motion-prior and tracking-based controllers have become a powerful foundation for physics-based character animation. Works like \citep{peng2021amp, tessler2024maskedmimic, xu2025parc, wu2025uniphys} learn reusable motor skills from large motion corpora, enabling a single controller to track diverse references, respond to different guidance signals, and produce natural long-horizon motions. However, these results are primarily demonstrated in simulation, where the controller can rely on accurate states, well-defined contacts, and resettable environments. Recent real-robot systems transfer this paradigm to humanoid hardware \citep{he2025asap,he2024omnih2o,li2025hold,liao2025beyondmimic,li2025amo,He2025LearningGP,Shao2025LangWBCLH,Xue2025AUA,Wang2025BeamDojoLA,ben2025homie,cheng2024expressive,li2025bfm,radosavovic2024humanoid,zhuang2024humanoid,zhang2025wococo,lau2026switch}. Methods like \citep{fu2024humanplus, ben2025homie, ze2025twist} build deployable whole-body trackers for shadowing, teleoperation, and data collection, while BeyondMimic~\citep{liao2025beyondmimic} further enhances the robustness and reliability of real-world deployments. SONIC~\citep{luo2025sonic} further scales motion tracking and synthesis to broad humanoid behaviors.
Utilizing motion-tracking controllers for teleoperation enables data collection and behavior cloning for a wide range of loco-manipulation tasks \citep{luo2025sonic,fu2024humanplus, shi2026egohumanoid}. However, this approach suffers from low data acquisition efficiency and sluggish motion response.

Applying reinforcement learning to loco-manipulation primarily falls into two categories. The first paradigm relies on task-specific reward engineering with customized goal formulations \citep{wang2025physhsi,lin2026lessmimic,he2025viral,xue2025opening, ren2025humanoid,zhang2026learning,su2025hitter}, such as box relocation \citep{wang2025physhsi}, opening doors \citep{xue2025opening}, or table tennis \citep{su2025hitter}. While these methods achieve strong performance on long-horizon tasks, they are inherently task-dependent and lack generalizability. Moreover, the control goals are usually sparse, further limiting control flexibility. The second paradigm extends motion tracking to Humanoid-Object Interaction (HOI) data, enabling unified control across diverse tasks via dense HOI interfaces \citep{yin2025visualmimic, zhao2025resmimic, zhang2025falcon, weng2025hdmi,yang2025omniretarget,wang2026humanx,he2026ultra}. Although these dense interfaces ensure high-fidelity execution by encoding detailed HOI coordination, the inherent complexity of HOI data renders them difficult to generate, edit, and use for online planning. Consequently, this complexity hinders the development of autonomous capabilities for long-horizon sequences. In contrast, we propose Contact Flow as a simple, flexible, and universal control interface for loco-manipulation. By extracting the physical essence of interaction, Contact Flow offers a representation that is significantly more compact and editable than dense HOI trajectories, yet more explicit than sparse goal states. It effectively bridges the gap by explicitly defining interaction-relevant body targets, object poses, and binary contact states for low-level controllers.

\subsection{Planning-Control Interfaces for Whole-Body Humanoid Control}
Long-horizon humanoid tasks necessitate a hierarchical approach, decoupling high-level task reasoning from low-level whole-body execution. Prior works bridge this gap using motion synthesis~\citep{xu2025parc,tessler2024maskedmimic,wu2025uniphys}, task-and-motion planning~\citep{ciebielski2025task,taouil2025physically,liu2025ego}, or vision-language models invoking skill libraries~\citep{xue2506leverb,schakkal2025hierarchical,jiang2025wholebodyvla}. However, the choice of the intermediate representation remains a critical bottleneck. Planners that output dense full-body trajectories are computationally prohibitive and inflexible for online replanning, whereas those outputting purely symbolic skills or sparse object goals deprive the executor of crucial contact guidance.

To implement this ideal intermediate representation, \textbf{Contact Flow}, \ours introduces a two-tier architecture. The high-level \textbf{CF-Gen} plans object-centric contact anchors and flow targets, while the low-level \textbf{CF-Track} executes them via a unified controller. By abstracting away dense kinematics while preserving essential contact semantics, this decoupled architecture uniquely enables real-time replanning and seamless skill composition.

\section{Method}
\label{sec:method}

\subsection{Overview}
Our goal is to enable humanoid robots to solve long-horizon loco-manipulation tasks by seamlessly chaining reusable meta-skills. The core challenge lies in designing an optimal skill interface: it must be expressive enough for contact-rich execution, yet compact enough for high-level planning.

\begin{figure*}[t]
    \centering
    \includegraphics[width=\linewidth]{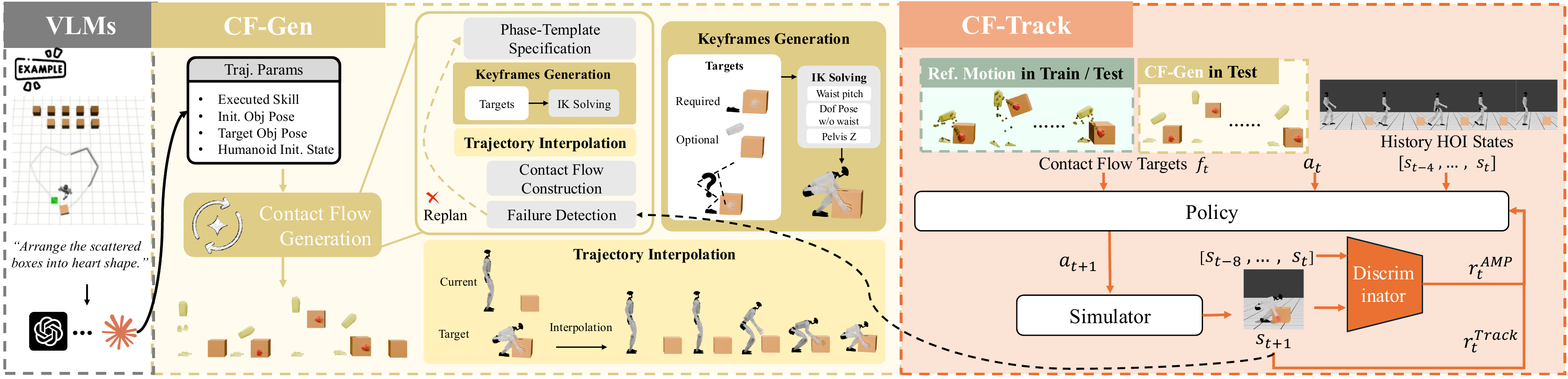}
    \caption{\textbf{Overview of \ours.} Given task goals and object states, CF-Gen heuristically synthesizes kinematic contact-flow segments, and CF-Track executes these segments through a robust low-level policy.}
    \label{fig:framework}
\end{figure*}

To this end, as illustrated in Fig.~\ref{fig:framework}, we propose \ours, a hierarchical framework centered on \textbf{contact flow} to explicitly decouple skill composition from skill execution. At the high-level, \textbf{CF-Gen} acts as a heuristic planner that synthesizes future contact-flow sequences, enabling real-time replanning for autonomous recovery. At the low-level, \textbf{CF-Track} serves as a unified executor, robustly realizing diverse, contact-rich loco-manipulation behaviors conditioned on the generated contact flow. By bridging high-level reasoning and low-level control through this shared representation, our method achieves seamless chaining of meta-skills for long-horizon tasks.

\begin{figure*}[t]
    \centering
    \includegraphics[width=\linewidth]{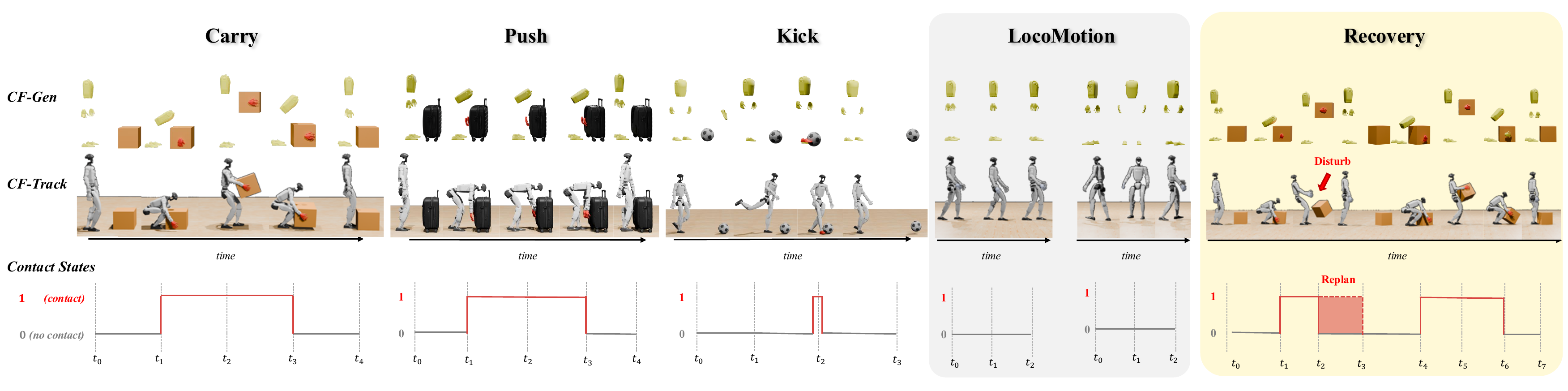}
    \caption{\textbf{Contact-flow examples across loco-manipulation skills.} Contact flow represents each skill with sparse future body targets and binary end-effector contact states, preserving interaction timing while avoiding dense whole-body trajectory commands.}
    \label{fig:contact_flow_examples}
\end{figure*}

\subsection{Contact Flow}
\label{sec:contact_flow}
We argue that the defining property of loco-manipulation is the active regulation of object contact dynamics. Motivated by this observation, we introduce \textbf{contact flow}, a compact intermediate representation designed to explicitly capture both whole-body motion intent and interaction structure.

Formally, the contact flow at each time step \(t\), denoted as \(\mathbf{F}_t\), is defined as a sequence of future interaction states. To capture both immediate and long-term intents, we non-uniformly sample future states at frame offsets \(\mathcal{T} = \{0, 1, 2, 3, 4, 8, 12, 16, 24, 32, 50\}\), the contact flow is formulated as:
\[
\mathbf{F}_t = \left\{ \left(\mathbf{b}_{t+k}, \mathbf{c}_{t+k}\right) \right\}_{k \in \mathcal{T}}
\]
where for each future step \(t+k\), \(\mathbf{b}_{t+k}\) denotes a sparse set of body-motion targets. \(\mathbf{c}_{t+k} \in \{0, 1\}^4\) is a 4-dimensional binary signal that explicitly specifies the contact states of the robot end-effectors. 

This formulation offers two critical advantages. First, it is \textbf{expressive}: as illustrated in Fig.~\ref{fig:contact_flow_examples}, it encodes diverse behaviors ranging from manipulation (e.g., carrying, pushing, kicking) to pure locomotion. Second, it is \textbf{compact and structured}: compared to dense, high-dimensional human-object interaction (HOI) trajectories, its sparse, non-uniform sampling makes it significantly easier to synthesize and sequence online. Together, these properties make contact flow an ideal interface, seamlessly bridging high-level heuristic planning and low-level robust control.

\subsection{CF-Track: Unified Execution of Loco-Manipulation Meta-Skills}
\label{sec:cf_track}
In our framework, CF-Track serves as the unified low-level executor, robustly realizing diverse loco-manipulation behaviors conditioned on the contact flow. Instead of learning separate, task-specific controllers, we train a unified policy tracking contact-flow targets across multiple interaction modes. 

\paragraph{Policy input and output.}
At each control step \(t\), the policy takes as input a comprehensive observation vector \(\mathbf{x}_t\), which concatenates the target contact flow \(\mathbf{F}_t\) (as defined in Sec.~\ref{sec:contact_flow}) and a history buffer of recent states \(\mathbf{H}_t\):
\[
\mathbf{x}_t = [ \mathbf{F}_t, \, \mathbf{H}_t ]
\]
The history buffer \(\mathbf{H}_t = [ \mathbf{o}_t, \dots, \mathbf{o}_{t-K+1} ]\) stores instantaneous observations over a window of \(K=5\) steps. Each observation \(\mathbf{o}_t = [ \mathbf{s}_t^{\text{prop}}, \mathbf{s}_t^{\text{obj}} ]\) comprises: (1) the proprioceptive state \(\mathbf{s}_t^{\text{prop}}\), including joint kinematics, base orientation, end-effector positions, and the previous action \(\mathbf{a}_{t-1}\); and (2) the object state \(\mathbf{s}_t^{\text{obj}}\), capturing its relative 6D pose and bounding box. Conditioned on \(\mathbf{x}_t\), the policy outputs low-level motor actions \(\mathbf{a}_t\) to drive the humanoid. See Tab.~\ref{tab:observation_terms} for details.

\paragraph{Learning objective.}
CF-Track is trained via reinforcement learning with following reward:
\[
r_t = \lambda_{\text{track}} r_t^{\text{track}} + \lambda_{\text{amp}} r_t^{\text{amp}} + \lambda_{\text{reg}} r_t^{\text{reg}},
\]
where \(r_t^{\text{track}}\) encourages tracking the target body and object trajectories, \(r_t^{\text{amp}}\) is an adversarial motion prior enforcing natural humanoid movements, and \(r_t^{\text{reg}}\) penalizes large action rates to ensure smooth control. See Appendix~\ref{sec:training_config} for details. Although both $r_t^{\mathrm{track}}$ and $r_t^{\mathrm{amp}}$ encourage data-consistent behavior, promote data-consistent behavior, they play complementary roles. Specifically, the tracking term defines the granularity of the target contact flow. An overly fine-grained target over-constrains the policy and limits the generalization, while an overly coarse style target provides insufficient guidance for accurate motion following. This trade-off motivates a more generalized contact-flow specification, as further analyzed in Tab.~\ref{tab:unified_cf_track_cf_gen_ablation}. 

\paragraph{Unified learning.}
Instead of learning task-specific skills, we train a single CF-Track policy on the OmniContact dataset (Appendix~\ref{sec:dataset}). Since contact flow compactly represents loco-manipulation, our policy unifies various interaction modes into a shared control. Despite being trained solely on human data, this robust formulation generalizes seamlessly during inference. It reliably tracks heuristically planned contact flows, translating imperfect target plans into stable humanoid execution.

\subsection{CF-Gen: Contact-Flow Synthesis and Skill Chaining}
\label{sec:cf_gen}

While CF-Track serves as a robust unified executor, long-horizon loco-manipulation further requires high-level decision-making to determine which meta-skill to invoke and how to instantiate its contact targets within the current scene. To bridge this gap, CF-Gen acts as a lightweight, rule-based reference synthesizer. Given an object-level goal, the current humanoid state, and the active object's pose and dimensions, CF-Gen generates a dense reference motion segment. This segment is subsequently converted online into a future contact flow, which is then consumed by CF-Track.

\paragraph{Phase-Template Specification.}
Rather than relying on computationally expensive full-body trajectory optimization, CF-Gen utilizes a compact library of hand-designed phase templates. Each meta-skill is decomposed into an ordered sequence of phase blocks, each characterized by explicit contact semantics. For instance, a carrying skill progresses sequentially through: approaching a pre-grasp stance, solving for a hand-contact pose, lifting the object, walking while maintaining contact, and finally releasing the object. For composed, long-horizon tasks, CF-Gen maintains a high-level stage state, systematically switching the active object, the applied meta-skill, and the target goal upon the completion of each stage. The complete phase-template library is detailed in Appendix~\ref{sec:phase_templates}.

\paragraph{Keyframes Generation.}
To adapt these templates to diverse scenes, CF-Gen anchors each phase by defining its ending pose through object-centric geometry. The specification of these target states is phase-dependent, allowing CF-Gen to selectively employ Inverse Kinematics (IK) only when precise end-effector placement is necessary. For purely locomotion-based phases, CF-Gen simply specifies the ending ankle poses based on the desired displacement, while the remaining joints assume a default nominal posture (\(\mathbf{q}_{\mathrm{default}}\)). Conversely, for phases establishing contact, CF-Gen determines the approach direction and selects contact anchors to specify both the ending ankle and wrist poses. The remaining posture is then resolved via a constrained IK problem. The optimizable variables include the pelvis height (\(z_{\mathrm{pelvis}}\)), pelvis pitch (\(\theta_{\mathrm{pitch}}\)), and all joint degrees of freedom (\(\mathbf{q}\)), while strictly excluding the waist roll and waist yaw to maintain torso stability. To ensure real-time synthesis, the IK is optimized for a maximum of 20 iterations by solving:
\begin{equation}
\small
\begin{aligned}
\Delta_e(\mathbf{q}, z, \theta)
&= \mathrm{FK}_e(\mathbf{q}, z, \theta) - \mathbf{x}_e^{\mathrm{tar}},\\
\mathcal{L}_{\mathrm{IK}}
&= \sum_{e \in \mathcal{E}}
\left\|\Delta_e(\mathbf{q}, z, \theta)\right\|^2\\
&\quad + \lambda \left\|\mathbf{q} - \mathbf{q}_{\mathrm{default}}\right\|^2,\\
\mathbf{q}^{\star}, z^{\star}_{\mathrm{pelvis}}, \theta^{\star}_{\mathrm{pitch}}
&= \arg\min_{\mathbf{q}, z, \theta} \mathcal{L}_{\mathrm{IK}}.
\end{aligned}
\end{equation}
where \(\mathcal{E}=\{\mathrm{wrists},\mathrm{ankles}\}\), \(\mathrm{FK}_e\) computes the forward kinematics for the specified end-effectors, and \(\lambda\) controls the regularization towards the default posture. 

\paragraph{Trajectory Interpolation.}
Following keyframe generation, CF-Gen synthesizes a continuous motion trajectory for each phase by interpolating between its start and ending poses. At this stage, the full kinematic state is represented by the pelvis position \(\mathbf{p} \in \mathbb{R}^3\), the pelvis orientation as a quaternion \(\mathbf{o} \in \mathbb{S}^3\), and the joint degrees of freedom \(\mathbf{q} \in \mathbb{R}^D\). Given the starting state and the target ending state for a specific phase of duration \(T\), we compute the intermediate state at any time \(t \in [0, T]\) using a normalized time parameter \(\alpha = t / T \in [0, 1]\). To properly handle the distinct geometric properties of these variables, we apply Linear Interpolation (LERP) for the Euclidean vectors and Spherical Linear Interpolation (SLERP) for the rotations:
$$
\begin{aligned}
\mathbf{p}(t) &= (1 - \alpha) \mathbf{p}_{\mathrm{start}} + \alpha \mathbf{p}_{\mathrm{end}}, \\
\mathbf{q}(t) &= (1 - \alpha) \mathbf{q}_{\mathrm{start}} + \alpha \mathbf{q}_{\mathrm{end}}, \\
\mathbf{o}(t) &= \mathrm{Slerp}(\mathbf{o}_{\mathrm{start}}, \mathbf{o}_{\mathrm{end}}, \alpha).
\end{aligned}
$$
This decoupled interpolation strategy ensures smooth transitions for translational movements and joint actuations, while the SLERP operation guarantees the shortest, constant-velocity rotational path for the pelvis. Consequently, this process bridges the discrete keyframes to yield a dense, full-body kinematic trajectory.

\paragraph{Contact Flow Construction.}
Rather than directly tracking the dense kinematic trajectory, CF-Track operates on a sparse, future-conditioned contact flow to maintain balance and compliance. At execution time, CF-Gen queries the interpolated dense reference at non-uniform future offsets \(\mathcal{T} = \{0, 1, 2, 3, 4, 8, 12, 16, 24, 32, 50\}\). For each offset \(\tau \in \mathcal{T}\), the target poses are transformed into the current torso-yaw frame to construct the contact flow:
$$
\mathbf{F}_t = \left\{ \left(\mathbf{b}_{t+\tau}, \mathbf{c}_{t+\tau}\right) \right\}_{\tau \in \mathcal{T}},
$$
where \(\mathbf{b}_{t+\tau}\) contains the sparse body targets (wrists, torso, ankles) and \(\mathbf{c}_{t+\tau} \in \{0,1\}^4\) denotes the binary contact states for the end-effectors. This formulation explicitly communicates the spatial and temporal intent of the contact without over-constraining the policy with dense, full-body joint commands, thereby preserving the humanoid's flexibility to execute natural movements.

\paragraph{Skill Chaining and Replanning.}
Long-horizon execution is achieved by seamlessly chaining these synthesized segments in a closed loop. To ensure robustness against perturbations, CF-Gen continuously monitors the execution at 50 Hz, detecting failures by comparing the observed and planned object states at the current time step \(t\):
$$
\delta_t = d\!\left(\mathbf{x}^{\text{obj}}_{t,\mathrm{obs}}, \mathbf{x}^{\text{obj}}_{t,\mathrm{pred}}\right).
$$
If the deviation \(\delta_t\) exceeds a predefined threshold \(\epsilon\)—due to unexpected events such as a dropped box or a missed contact—CF-Gen immediately aborts the current plan and replans from the current state. This high-frequency feedback loop naturally elicits reactive recovery behaviors, enabling the humanoid to re-approach the object and resume the task autonomously.

\subsection{Hierarchical Execution}
\label{sec:hierarchical_execution}
\begin{figure*}[t]
    \centering
    \includegraphics[width=0.98\linewidth]{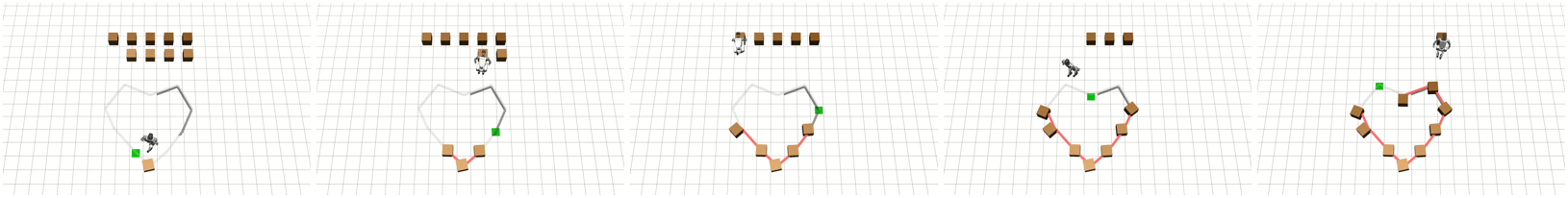}
    \caption{\textbf{VLM integration examples.} Given prompt: \textit{``Arrange scattered boxes into a heart shape.''}}
    \label{fig:extra_vlm_examples_main}
\end{figure*}

At test time, \ours operates as a hierarchical closed-loop system (Fig.~\ref{fig:framework}). CF-Gen translates a task goal into object-centric contact-flow segments, which CF-Track executes while continuously monitoring the robot and object states. Upon phase completion or failure detection, CF-Gen updates the state and synthesizes the next segment. This cycle repeats until the task is completed. Furthermore, as shown in Fig.~\ref{fig:extra_vlm_examples_main}, \ours supports VLM integration for more complex tasks, including language-grounded transfer and concept-driven layout. See Appendix~\ref{sec:vlm_integration} for details.

\section{Experiments}
\label{sec:experiment}

\begin{table*}[t]
    \centering
    \captionsetup{skip=3pt}
    \small
    \caption{\textbf{Benchmarked comparison in simulation}. ``--'' denotes unsupported tasks. For Meta-Skill Chaining tasks, all reported numbers in different stages represent the success rate $R_{\text{succ}}(\%)$.}
    \label{tab:simulation_benchmark}
    \setlength{\tabcolsep}{3.5pt}
    \renewcommand{\arraystretch}{1.15}
    \resizebox{\textwidth}{!}{
        \begin{tabular}{l c c c c c c c c c c c}
            \toprule
            \multirow{4}{*}{Methods} & \multicolumn{6}{c}{Meta-Skill} & \multicolumn{5}{c}{Meta-Skill Chaining} \\
            \cmidrule(lr){2-7} \cmidrule(lr){8-12}
            & \multicolumn{3}{c}{\textit{Carry Box}} & \multicolumn{3}{c}{\textit{Push Suitcase}} & \multicolumn{3}{c}{\textit{Stack Boxes} } & \multicolumn{2}{c}{\textit{Push-Stack Boxes}} \\
            \cmidrule(lr){2-4} \cmidrule(lr){5-7} \cmidrule(lr){8-10} \cmidrule(lr){11-12}
            & $R_{\text{succ}}(\%)\uparrow$ & $E_{\text{obj}}^{T}\downarrow$ & $N_{\text{hoi}}\uparrow$
            & $R_{\text{succ}}(\%)\uparrow$ & $E_{\text{obj}}^{T}\downarrow$ & $N_{\text{hoi}}\uparrow$
            & Stage 1 & Stage 2 & Stage 3
            & Stage 1 & Stage 2 \\
            \midrule
            Sonic~\cite{luo2025sonic} & 
            $3.38^{\textcolor{gray}{(\pm 1.5)}}$ & $5.21^{\textcolor{gray}{(\pm 0.1)}}$ & $1.25^{\textcolor{gray}{(\pm 0.4)}}$
            & $0.00^{\textcolor{gray}{(\pm 0.0)}}$ & $4.96^{\textcolor{gray}{(\pm 0.1)}}$ & $1.14^{\textcolor{gray}{(\pm 0.4)}}$
            & $0.0$ & $0.0$ & $0.0$ & $0.0$ & $0.0$ \\
            HDMI~\cite{weng2025hdmi}
            & $0.00^{\textcolor{gray}{(\pm 0.0)}}$ & $5.35^{\textcolor{gray}{(\pm 0.1)}}$ & $0.00^{\textcolor{gray}{(\pm 0.0)}}$
            & $0.00^{\textcolor{gray}{(\pm 0.0)}}$ & $5.11^{\textcolor{gray}{(\pm 0.3)}}$ & $0.00^{\textcolor{gray}{(\pm 0.0)}}$
            & $0.0$ & $0.0$ & $0.0$ & $0.0$ & $0.0$ \\
            PhysHSI~\cite{wang2025physhsi}
            & $\underline{87.00}^{\textcolor{gray}{(\pm 2.4)}}$ & $\underline{0.58}^{\textcolor{gray}{(\pm 0.1)}}$ & $\underline{6.62}^{\textcolor{gray}{(\pm 2.2)}}$
            & -- & -- & --
            & $\underline{82.0}$ & $\underline{56.5}$ & $0.0$ & -- & -- \\
            LessMimic~\cite{lin2026lessmimic}
            & $34.00^{\textcolor{gray}{(\pm 3.4)}}$ & $2.60^{\textcolor{gray}{(\pm 0.2)}}$ & $3.24^{\textcolor{gray}{(\pm 1.1)}}$
            & $\underline{12.50}^{\textcolor{gray}{(\pm 2.4)}}$ & $\underline{3.14}^{\textcolor{gray}{(\pm 0.1)}}$ & $\underline{1.86}^{\textcolor{gray}{(\pm 1.1)}}$
            & $21.0$ & $3.5$ & $0.0$ & $\underline{9.0}$ & $0.0$ \\
            \textbf{OmniContact}
            & $\textbf{98.70}^{\textcolor{gray}{(\pm 0.6)}}$ & $\textbf{0.07}^{\textcolor{gray}{(\pm 0.0)}}$ & $\textbf{7.75}^{\textcolor{gray}{(\pm 0.4)}}$
            & $\textbf{82.50}^{\textcolor{gray}{(\pm 2.7)}}$ & $\textbf{0.27}^{\textcolor{gray}{(\pm 0.0)}}$ & $\textbf{6.00}^{\textcolor{gray}{(\pm 0.5)}}$
            & $\textbf{89.0}$ & $\textbf{87.0}$ & $\textbf{56.5}$ & $\textbf{91.5}$ & $\textbf{76.5}$ \\
            \bottomrule
        \end{tabular}}
\end{table*}

\subsection{Experimental Setup}
\paragraph{Tasks.}
We report the evaluation of \ours on four representative humanoid loco-manipulation tasks. For meta-skills, \textit{Carry Box} involves lifting and transporting a box to a target, and \textit{Push Suitcase} requires aligning and pushing a suitcase to a goal. To evaluate sequential skill chaining, \textit{Stack Boxes} entails gathering and stacking three scattered boxes. Finally, for skill composition, \textit{Push-Stack Boxes} combines pushing a suitcase and stacking a box atop it. Appendix~\ref{sec:append_task} details these tasks, additional meta-skills (e.g., \textit{Slide Box}, \textit{Kick Ball}), and further chaining scenarios.

\paragraph{Evaluation metrics.}
Our primary metric is the task success rate \(R_{\text{succ}}\), evaluated across randomized configurations (Appendix~\ref{sec:append_eval_protocol}), with \(R_{\text{succ}}^{*}\) specifically denoting the success rate under online replanning. We also report the final object error \(E_{\text{obj}}^{T}\) for meta-skills. To assess motion quality, we introduce \(N_{\text{hoi}}\), a video-based naturalness score evaluating stability, contact plausibility, and smoothness (Appendix~\ref{sec:append_naturalness_eval}). Each evaluation uses five random seeds, with 200 randomized initial-final goal pairs per seed. We report the average value and standard deviation across seeds. Comprehensive ablations on overall average tracking errors (\(E_{\text{torso}}, E_{\text{obj}}\)) and success rates are conducted to validate our core design choices, including the tracking target, synthesis configurations, and reward balancing. Unless otherwise specified, \ours is trained on the OmniContact dataset, our self-collected MoCap-based HOI corpus for humanoid loco-manipulation. See details in Appendix~\ref{sec:dataset}.

\paragraph{Baselines}
We compare against motion tracking (Sonic~\cite{luo2025sonic}) and interaction learning (HDMI~\cite{weng2025hdmi}, PhysHSI~\cite{wang2025physhsi}, LessMimic~\cite{lin2026lessmimic}) baselines under identical randomized conditions, with necessary adaptations. Specifically, LessMimic is evaluated only on the XY plane for \textit{Carry Box} (lacking height-release capability) and initialized with objects in-hand for pushing tasks (lacking autonomous approach). For Sonic and HDMI, we provide the required dense tracking references: we use MoCap-retargeted data when available and synthesize dense references from the task metadata otherwise. See Appendix~\ref{sec:append_eval_protocol} for protocol details and a meta-skill fairness subset.
The appendix subset is a mean-only diagnostic evaluation constructed from a pure MoCap-data subset, and is used to isolate controller-level fairness under matched initial poses, waypoints, and target poses; its numbers are therefore not intended to replace the full randomized benchmark in Table~\ref{tab:simulation_benchmark}.

\subsection{Overall Performance}
\label{sec:main_eval}
\paragraph{Base Performance} Table~\ref{tab:simulation_benchmark} summarizes the simulation benchmark, where \ours demonstrates superior performance across all tasks. We focus our primary evaluation on contact-rich HOI tasks, deferring basic locomotion metrics to Appendix~\ref{sec:append_locomotion_eval}. We highlight two key findings: (1) \textbf{High single-skill success.} Our method achieves dominant success rates on \textit{Carry Box} (\(98.7\%\)) and \textit{Push Suitcase} (\(82.5\%\)). Notably, while the motion-tracking baseline Sonic~\cite{luo2025sonic} smoothly tracks walking, it exhibits severe instability during bending or squatting and fails to lift objects. Similarly, HDMI fails immediately because its single-trajectory policy overfits the training states and cannot generalize to randomized test states. These failures highlight that relying solely on body kinematics or narrow trajectory memorization is insufficient for robust HOI.
(2) \textbf{Long-horizon composability.} \ours successfully solves multi-stage tasks, whereas all baselines completely fail (\(0\%\)) due to fragile long-horizon execution (\textit{Stack Boxes}) or missing skill transitions (\textit{Push-Stack Boxes}).

\paragraph{Online Replanning} Table~\ref{tab:closed_loop_recovery} evaluates the impact of online replanning. This dynamic adjustment consistently boosts success rates, notably raising \textit{Push Suitcase} to \(94.5\%\) and \textit{Stack Boxes} to \(80.5\%\). We further observe that emergency recoveries are rarely triggered, and the performance gains primarily stem from refreshing subsequent segments with updated object states.

\begin{table}[t]
    \centering
    \small
    \caption{\textbf{Results with online replanning}.}
    \label{tab:closed_loop_recovery}
    \setlength{\tabcolsep}{4pt}
    \renewcommand{\arraystretch}{1.2}
    \resizebox{\linewidth}{!}{
    \begin{tabular}{lccc}
        \toprule
        Task & \(R_{\text{succ}}\) (\%) & \(R_{\text{succ}}^{*}\) (\%) & Avg. Replans \\
        \midrule
        \textit{Carry Box} & 98.7 & 99.7 & 0.01 \\
        \textit{Push Suitcase} & 82.5 & 94.5 & 0.77 \\
        \textit{Stack Boxes} & 56.6 & 80.5 & 0.96 \\
        \textit{Push-Stack Boxes} & 76.5 & 84.5 & 0.84 \\
        \bottomrule
    \end{tabular}}
\end{table}

\begin{figure}[t]
    \centering
    \includegraphics[width=\linewidth]{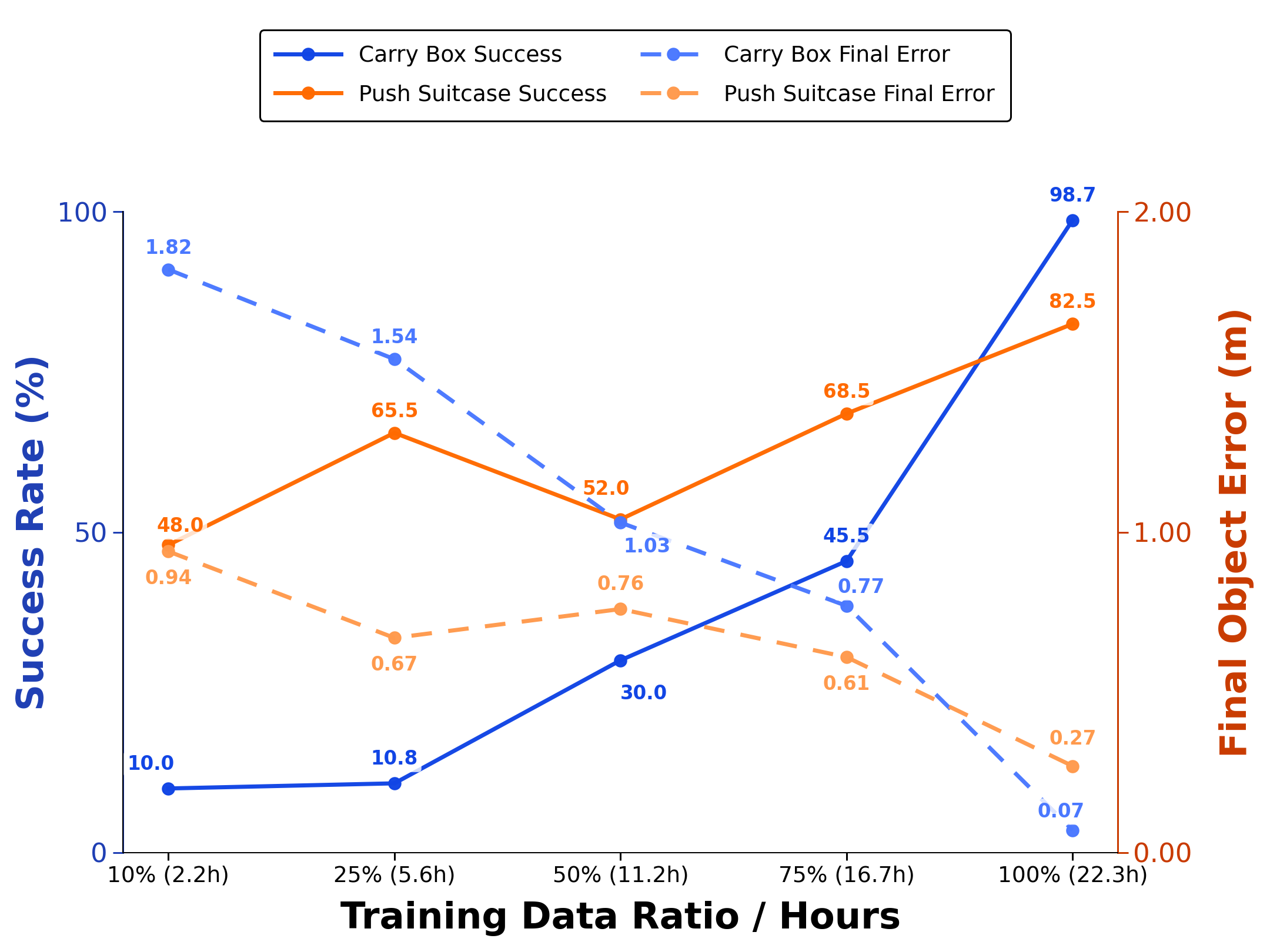}
    \caption{\textbf{Scaling with HOI data size.}}
    \label{fig:scalability_ablation}
\end{figure}

\begin{table}[t]
    \centering
    \small
    \caption{\textbf{\textcolor{red}{[Contact Flow]} Ablation on tracking target design.} See Sec.~\ref{sec:ablate_eval} for detailed explanation.}
    \label{tab:cf_track_tracking_target_ablation}
    \setlength{\tabcolsep}{4pt}
    \renewcommand{\arraystretch}{1.15}
    \resizebox{\linewidth}{!}{
    \begin{tabular}{l | c c | c c c}
        \toprule
        \multirow{2}{*}{Tracking Targets} & \multicolumn{2}{c|}{Training / Test Set} & \multicolumn{3}{c}{CF-Gen Meta-Skill} \\
        \cmidrule(lr){2-3} \cmidrule(lr){4-6}
        & $E_{\text{torso}}\downarrow$ & $E_{\text{obj}}\downarrow$ & $E_{\text{torso}}\downarrow$ & $E_{\text{obj}}\downarrow$ & $R_{\text{succ}}(\%)\uparrow$ \\
        \midrule
        Torso only & 0.12 / 0.14 & 0.48 / 0.54 & \underline{0.18} & 1.83 & 0.50 \\
        {[T, EE]} & \textbf{0.10} / \textbf{0.10} & \underline{0.36} / 0.45 & 0.22 & 1.53 & 11.50 \\
        {[T, EE, O]} & \underline{0.11} / \underline{0.12} & 0.41 / 0.49 & 0.24 & 1.29 & 6.50 \\
        {[T, EE, C, O]} & 0.12 / 0.14 & 0.37 / \underline{0.44} & 0.20 & \underline{0.72} & \underline{22.50} \\
        {[T, EE, C, O, Dof]} & 0.15 / 0.18 & 0.54 / 0.63 & 1.49 & 1.92 & 0.00 \\
        {[T, FB, C, O, Dof]} & \textbf{0.10} / 0.18 & \textbf{0.28} / \textbf{0.40} & 0.74 & 1.89 & 0.50 \\
        \midrule
        {[T, EE, C]} (Ours) & \textbf{0.10} / \textbf{0.10} & 0.43 / 0.45 & \textbf{0.13} & \textbf{0.15} & \textbf{98.70} \\
        \bottomrule
    \end{tabular}}
\end{table}

\begin{table}[t]
    \centering
    \small
    \caption{\textbf{Ablation of CF-Gen and CF-Track.}}
    \label{tab:unified_cf_track_cf_gen_ablation}
    \setlength{\tabcolsep}{4pt}
    \renewcommand{\arraystretch}{1.08}
    \resizebox{\linewidth}{!}{
    \begin{tabular}{l c c c c | c}
        \toprule
        \rowcolor{gray!15}
        \multicolumn{6}{l}{\textit{\textbf{\textcolor{orange}{[CF-Gen]} Synthesis Configuration}}}\\
        \midrule
        Metric & \shortstack{w/o contact\\adapt.} & \shortstack{w/o torso\\adapt.} & \shortstack{w/o wrist\\adapt.} & \shortstack{w/o\\replan} & \shortstack{Full CF-Gen\\(Ours)} \\
        \midrule
        $E_{\text{obj}}^T\downarrow$ & 0.14 & 0.70 & 0.14 & \underline{0.07} & \textbf{0.05} \\
        $R_{\text{succ}}(\%)\uparrow$ & 84.3 & 77.0 & 96.9 & \underline{98.70} & \textbf{99.70} \\
        \midrule
        \rowcolor{gray!15}
        \multicolumn{6}{l}{\textit{\textbf{\textcolor{darkred}{[CF-Track]} Reward Balance ($W_{\text{track}}$-$W_{\text{amp}}$)}}}\\
        \midrule
        Metric & 0.3--0.7 & 0.5--0.5 & 0.7--0.3 & 1.0--0.0 & 0.85--0.15 (Ours) \\
        \midrule
        $E_{\text{torso}}\downarrow$ & 1.32 & 1.40 & 0.34 & \textbf{0.08} & \underline{0.12} \\
        $E_{\text{obj}}\downarrow$ & 0.26 & 0.17 & 0.15 & \underline{0.13} & \textbf{0.12} \\
        $R_{\text{succ}}(\%)\uparrow$ & 1.60 & 0.10 & 53.70 & \underline{88.90} & \textbf{98.70} \\
        $R_{\text{stable}}(\%)\uparrow$ & \underline{73.60} & \textbf{82.70} & 71.60 & 46.30 & 58.80 \\
        \bottomrule
    \end{tabular}}
\end{table}

\subsection{Ablation Study}
\label{sec:ablate_eval}

\paragraph{\textcolor{red}{[Contact Flow]} Tracking target design.} 
Table~\ref{tab:cf_track_tracking_target_ablation} evaluates contact flow targets. Torso-only (T) tracking is too sparse for interaction, while dense targets (O, Dof, FB) overconstrain the policy. These excessive constraints limit the policy's flexibility, resulting in CF-Gen tracking task failures. Crucially, explicitly modeling contact (C) provides necessary guidance, boosting the [T, EE] baseline's success from \(11.50\%\) to \(98.70\%\) while maintaining minimal errors. Ultimately, our [T, EE, C] interface optimally balances proper constraint for intention learning with compactness for CF-Gen.

\paragraph{\textcolor{orange}{[CF-Gen]} Synthesis configuration.} 
Table~\ref{tab:unified_cf_track_cf_gen_ablation} ablates the components of the CF-Gen trajectory synthesis pipeline on the \textit{Carry Box} task. CF-Gen plans object-centric motion by selecting the optimal contact face, adapting torso and wrist targets to match object geometry, and replanning online when execution deviates. As shown, removing any of these modules increases $E^T_{\text{obj}}$ and reduces $R_{\text{succ}}$. 

\paragraph{\textcolor{darkred}{[CF-Track]} Reward balance.}
Table~\ref{tab:unified_cf_track_cf_gen_ablation} ablates the trade-off between tracking (\(W_{\text{track}}\)) and the adversarial motion prior (\(W_{\text{amp}}\)). Low tracking weights (0.5--0.5) over-prioritize motion naturalness, yielding high stability (\(82.70\%\)) but near-zero task success. Conversely, pure tracking minimizes torso error but weakens robustness against disturbances, dropping stability to \(46.30\%\). Crucially, our selected balance preserves motion-prior regularization without overriding tracking, achieving peak task success (\(98.70\%\)) alongside minimal object error. This balance lets CF-Track follow CF-Gen targets while smoothing rule-based trajectory artifacts, leading to the highest overall success.

\paragraph{Scalability with data volume.}
Fig.~\ref{fig:scalability_ablation} shows that scaling HOI data from 10\%-2.2h to 100\%-22.3h of the OmniContact dataset (Appendix~\ref{sec:dataset}) improves both success and object accuracy. This strong scaling behavior shows that \ours effectively captures diverse data distributions, highlighting its promising potential as a robust, universal foundation for HOI tracking.

\section{Conclusion and Discussion}
\label{sec:conclusion}

We introduced \ours, a hierarchical framework that leverages \textbf{contact flow} to bridge the gap between high-level task reasoning and low-level whole-body execution. By unifying the system through this compact interface—where CF-Track handles execution and CF-Gen manages replanning—our approach supports robust skill chaining and seamless VLM integration. Together with the OmniContact dataset that we collect for this problem setting, \ours demonstrates that decoupling interaction semantics (\emph{what}) from physical execution (\emph{how}) is key to scalable loco-manipulation.
Nevertheless, current limitations suggest clear directions for future work. First, our underactuated grippers limit fine manipulation; extending contact flow to dexterous hands is a key priority. Second, while effective, our rule-based CF-Gen planner struggles with highly dynamic scenarios. We envision replacing these heuristics with a learnable, data-driven approach for generating contact anchors. 
Crucially, since contact flow abstracts away the complexity of full-body dynamics, it serves as an ideal signal for learning from in-the-wild human videos, thereby unlocking the potential for highly reactive and natural humanoid behaviors.

\section{Acknowledgments}
\label{sec:acknowledgments}
We would like to thank Hanyang Cao for his invaluable assistance with motion retargeting. We thank Tao Huang and Qihan Zhao for their help in setting up the motion capture system. We are also grateful to Haonan Zhang for his technical support with 3D printing. Finally, we extend our deep appreciation to the Noitom Robotics data collection team and the motion capture actors for their dedication, cooperation, and constructive feedback throughout the data collection and system refinement phases.

\bibliography{omnicontact}

@article{he2025asap,
  title={ASAP: Aligning Simulation and Real-World Physics for Learning Agile Humanoid Whole-Body Skills},
  author={He, Tairan and Gao, Jiawei and Xiao, Wenli and Zhang, Yuanhang and Wang, Zi and Wang, Jiashun and Luo, Zhengyi and He, Guanqi and Sobanbab, Nikhil and Pan, Chaoyi and others},
  journal={arXiv preprint arXiv:2502.01143},
  year={2025}
}

@article{wang2025physhsi,
  title={Physhsi: Towards a real-world generalizable and natural humanoid-scene interaction system},
  author={Wang, Huayi and Zhang, Wentao and Yu, Runyi and Huang, Tao and Ren, Junli and Jia, Feiyu and Wang, Zirui and Niu, Xiaojie and Chen, Xiao and Chen, Jiahe and others},
  journal={arXiv preprint arXiv:2510.11072},
  year={2025}
}

@article{lin2026lessmimic,
  title={Lessmimic: Long-horizon humanoid interaction with unified distance field representations},
  author={Lin, Yutang and Cui, Jieming and Li, Yixuan and Jia, Baoxiong and Zhu, Yixin and Huang, Siyuan},
  journal={arXiv preprint arXiv:2602.21723},
  year={2026}
}

@inproceedings{li2025hold,
  title={Hold My Beer: Learning Gentle Humanoid Locomotion and End-Effector Stabilization Control},
  author={Li, Yitang and Zhang, Yuanhang and Xiao, Wenli and Pan, Chaoyi and Weng, Haoyang and He, Guanqi and He, Tairan and Shi, Guanya},
  booktitle={RSS 2025 Workshop on Whole-body Control and Bimanual Manipulation: Applications in Humanoids and Beyond},
  year={2025}
}

@article{Shao2025LangWBCLH,
  title={LangWBC: Language-directed Humanoid Whole-Body Control via End-to-end Learning},
  author={Yiyang Shao and Xiaoyu Huang and Bike Zhang and Qiayuan Liao and Yuman Gao and Yufeng Chi and Zhongyu Li and Sophia Shao and Koushil Sreenath},
  journal={ArXiv},
  year={2025},
  volume={abs/2504.21738},
}

@article{li2025bfm,
  title={Bfm-zero: A promptable behavioral foundation model for humanoid control using unsupervised reinforcement learning},
  author={Li, Yitang and Luo, Zhengyi and Zhang, Tonghe and Dai, Cunxi and Kanervisto, Anssi and Tirinzoni, Andrea and Weng, Haoyang and Kitani, Kris and Guzek, Mateusz and Touati, Ahmed and others},
  journal={arXiv preprint arXiv:2511.04131},
  year={2025}
}

@article{cheng2024expressive,
  title={Expressive Whole-Body Control for Humanoid Robots},
  author={Cheng, Xuxin and Ji, Yandong and Chen, Junming and Yang, Ruihan and Yang, Ge and Wang, Xiaolong},
  journal={arXiv preprint arXiv:2402.16796},
  year={2024}
}

@article{radosavovic2024humanoid,
  title={Humanoid locomotion as next token prediction},
  author={Radosavovic, Ilija and Zhang, Bike and Shi, Baifeng and Rajasegaran, Jathushan and Kamat, Sarthak and Darrell, Trevor and Sreenath, Koushil and Malik, Jitendra},
  journal={Advances in neural information processing systems},
  volume={37},
  pages={79307--79324},
  year={2024}
}

@inproceedings{zhang2025wococo,
  title={WoCoCo: Learning Whole-Body Humanoid Control with Sequential Contacts},
  author={Zhang, Chong and Xiao, Wenli and He, Tairan and Shi, Guanya},
  booktitle={Conference on Robot Learning},
  pages={455--472},
  year={2025},
  organization={PMLR}
}

@article{zhuang2024humanoid,
  title={Humanoid Parkour Learning},
  author={Zhuang, Ziwen and Yao, Shenzhe and Zhao, Hang},
  journal={arXiv preprint arXiv:2406.10759},
  year={2024}
}

@article{Wang2025BeamDojoLA,
  title={BeamDojo: Learning Agile Humanoid Locomotion on Sparse Footholds},
  author={Huayi Wang and Zirui Wang and Junli Ren and Qingwei Ben and Tao Huang and Weinan Zhang and Jiangmiao Pang},
  journal={ArXiv},
  year={2025},
  volume={abs/2502.10363},
}

@article{Xue2025AUA,
  title={A Unified and General Humanoid Whole-Body Controller for Fine-Grained Locomotion},
  author={Yufei Xue and Wentao Dong and Minghuan Liu and Weinan Zhang and Jiangmiao Pang},
  journal={ArXiv},
  year={2025},
  volume={abs/2502.03206},
}

@article{He2025LearningGP,
  title={Learning Getting-Up Policies for Real-World Humanoid Robots},
  author={Xialin He and Runpei Dong and Zixuan Chen and Saurabh Gupta},
  journal={ArXiv},
  year={2025},
  volume={abs/2502.12152},
}

@article{li2025amo,
  title={AMO: Adaptive Motion Optimization for Hyper-Dexterous Humanoid Whole-Body Control},
  author={Li, Jialong and Cheng, Xuxin and Huang, Tianshu and Yang, Shiqi and Qiu, Ri-Zhao and Wang, Xiaolong},
  journal={arXiv preprint arXiv:2505.03738},
  year={2025}
}

@article{weng2025hdmi,
  title={Hdmi: Learning interactive humanoid whole-body control from human videos},
  author={Weng, Haoyang and Li, Yitang and Sobanbabu, Nikhil and Wang, Zihan and Luo, Zhengyi and He, Tairan and Ramanan, Deva and Shi, Guanya},
  journal={arXiv preprint arXiv:2509.16757},
  year={2025}
}

@article{yang2025omniretarget,
  title={Omniretarget: Interaction-preserving data generation for humanoid whole-body loco-manipulation and scene interaction},
  author={Yang, Lujie and Huang, Xiaoyu and Wu, Zhen and Kanazawa, Angjoo and Abbeel, Pieter and Sferrazza, Carmelo and Liu, C Karen and Duan, Rocky and Shi, Guanya},
  journal={arXiv preprint arXiv:2509.26633},
  year={2025}
}

@article{he2026ultra,
  title={ULTRA: Unified Multimodal Control for Autonomous Humanoid Whole-Body Loco-Manipulation},
  author={He, Xialin and Xu, Sirui and Li, Xinyao and Dong, Runpei and Bian, Liuyu and Wang, Yu-Xiong and Gui, Liang-Yan},
  journal={arXiv preprint arXiv:2603.03279},
  year={2026}
}

@article{peng2021amp,
  title={Amp: Adversarial motion priors for stylized physics-based character control},
  author={Peng, Xue Bin and Ma, Ze and Abbeel, Pieter and Levine, Sergey and Kanazawa, Angjoo},
  journal={ACM Transactions on Graphics (ToG)},
  volume={40},
  number={4},
  pages={1--20},
  year={2021},
  publisher={ACM New York, NY, USA}
}

@article{tessler2024maskedmimic,
  title={Maskedmimic: Unified physics-based character control through masked motion inpainting},
  author={Tessler, Chen and Guo, Yunrong and Nabati, Ofir and Chechik, Gal and Peng, Xue Bin},
  journal={ACM Transactions On Graphics (TOG)},
  volume={43},
  number={6},
  pages={1--21},
  year={2024},
  publisher={ACM New York, NY, USA}
}

@inproceedings{xu2025parc,
  title={Parc: Physics-based augmentation with reinforcement learning for character controllers},
  author={Xu, Michael and Shi, Yi and Yin, KangKang and Peng, Xue Bin},
  booktitle={Proceedings of the Special Interest Group on Computer Graphics and Interactive Techniques Conference Conference Papers},
  pages={1--11},
  year={2025}
}

@inproceedings{wu2025uniphys,
  title={Uniphys: Unified planner and controller with diffusion for flexible physics-based character control},
  author={Wu, Yan and Karunratanakul, Korrawe and Luo, Zhengyi and Tang, Siyu},
  booktitle={Proceedings of the IEEE/CVF International Conference on Computer Vision},
  pages={13214--13224},
  year={2025}
}

@article{ben2025homie,
  title={Homie: Humanoid loco-manipulation with isomorphic exoskeleton cockpit},
  author={Ben, Qingwei and Jia, Feiyu and Zeng, Jia and Dong, Junting and Lin, Dahua and Pang, Jiangmiao},
  journal={arXiv preprint arXiv:2502.13013},
  year={2025}
}

@article{liao2025beyondmimic,
  title={Beyondmimic: From motion tracking to versatile humanoid control via guided diffusion},
  author={Liao, Qiayuan and Truong, Takara E and Huang, Xiaoyu and Gao, Yuman and Tevet, Guy and Sreenath, Koushil and Liu, C Karen},
  journal={arXiv preprint arXiv:2508.08241},
  year={2025}
}

@article{luo2025sonic,
  title={Sonic: Supersizing motion tracking for natural humanoid whole-body control},
  author={Luo, Zhengyi and Yuan, Ye and Wang, Tingwu and Li, Chenran and Chen, Sirui and Castaneda, Fernando and Cao, Zi-Ang and Li, Jiefeng and Minor, David and Ben, Qingwei and others},
  journal={arXiv preprint arXiv:2511.07820},
  year={2025}
}

@article{yin2025visualmimic,
  title={Visualmimic: Visual humanoid loco-manipulation via motion tracking and generation},
  author={Yin, Shaofeng and Ze, Yanjie and Yu, Hong-Xing and Liu, C Karen and Wu, Jiajun},
  journal={arXiv preprint arXiv:2509.20322},
  year={2025}
}

@article{zhao2025resmimic,
  title={Resmimic: From general motion tracking to humanoid whole-body loco-manipulation via residual learning},
  author={Zhao, Siheng and Ze, Yanjie and Wang, Yue and Liu, C Karen and Abbeel, Pieter and Shi, Guanya and Duan, Rocky},
  journal={arXiv preprint arXiv:2510.05070},
  year={2025}
}

@article{zhang2025falcon,
  title={Falcon: Learning force-adaptive humanoid loco-manipulation},
  author={Zhang, Yuanhang and Yuan, Yifu and Gurunath, Prajwal and Gupta, Ishita and Omidshafiei, Shayegan and Agha-mohammadi, Ali-akbar and Vazquez-Chanlatte, Marcell and Pedersen, Liam and He, Tairan and Shi, Guanya},
  journal={arXiv preprint arXiv:2505.06776},
  year={2025}
}

@article{wang2026humanx,
  title={HumanX: Toward Agile and Generalizable Humanoid Interaction Skills from Human Videos},
  author={Wang, Yinhuai and Zhao, Qihan and Lau, Yuen Fui and Yu, Runyi and Tsui, Hok Wai and Chen, Qifeng and Wang, Jingbo and Pang, Jiangmiao and Tan, Ping},
  journal={arXiv preprint arXiv:2602.02473},
  year={2026}
}

@inproceedings{ciebielski2025task,
  title={Task and Motion Planning for Humanoid Loco-Manipulation},
  author={Ciebielski, Michal and Dh{\'e}din, Victor and Khadiv, Majid},
  booktitle={2025 IEEE-RAS 24th International Conference on Humanoid Robots (Humanoids)},
  pages={1179--1186},
  year={2025},
  organization={IEEE}
}

@inproceedings{taouil2025physically,
  title={Physically consistent humanoid loco-manipulation using latent diffusion models},
  author={Taouil, Ilyass and Zhao, Haizhou and Dai, Angela and Khadiv, Majid},
  booktitle={2025 IEEE-RAS 24th International Conference on Humanoid Robots (Humanoids)},
  pages={1--8},
  year={2025},
  organization={IEEE}
}

@article{liu2025ego,
  title={Ego-Vision World Model for Humanoid Contact Planning},
  author={Liu, Hang and Gao, Yuman and Teng, Sangli and Chi, Yufeng and Shao, Yakun Sophia and Li, Zhongyu and Ghaffari, Maani and Sreenath, Koushil},
  journal={arXiv preprint arXiv:2510.11682},
  year={2025}
}

@article{xue2506leverb,
  title={LeVERB: Humanoid Whole-Body Control with Latent Vision-Language Instruction,(2025)},
  author={Xue, Haoru and Huang, Xiaoyu and Niu, Dantong and Liao, Qiayuan and Kragerud, Thomas and Gravdahl, Jan Tommy and Peng, Xue Bin and Shi, Guanya and Darrell, Trevor and Sreenath, Koushil and others},
  journal={URL https://arxiv. org/abs/2506.13751},
  volume={3},
  number={10},
  year={2025}
}

@article{schakkal2025hierarchical,
  title={Hierarchical vision-language planning for multi-step humanoid manipulation},
  author={Schakkal, Andr{\'e} and Zandonati, Ben and Yang, Zhutian and Azizan, Navid},
  journal={arXiv preprint arXiv:2506.22827},
  year={2025}
}

@article{jiang2025wholebodyvla,
  title={Wholebodyvla: Towards unified latent vla for whole-body loco-manipulation control},
  author={Jiang, Haoran and Chen, Jin and Bu, Qingwen and Chen, Li and Shi, Modi and Zhang, Yanjie and Li, Delong and Suo, Chuanzhe and Wang, Chuang and Peng, Zhihui and others},
  journal={arXiv preprint arXiv:2512.11047},
  year={2025}
}

@article{ze2025twist,
  title={Twist: Teleoperated whole-body imitation system},
  author={Ze, Yanjie and Chen, Zixuan and Ara{\'u}jo, Joao Pedro and Cao, Zi-ang and Peng, Xue Bin and Wu, Jiajun and Liu, C Karen},
  journal={arXiv preprint arXiv:2505.02833},
  year={2025}
}

@article{fu2024humanplus,
  title={Humanplus: Humanoid shadowing and imitation from humans},
  author={Fu, Zipeng and Zhao, Qingqing and Wu, Qi and Wetzstein, Gordon and Finn, Chelsea},
  journal={arXiv preprint arXiv:2406.10454},
  year={2024}
}

@article{shi2026egohumanoid,
  title={Egohumanoid: Unlocking in-the-wild loco-manipulation with robot-free egocentric demonstration},
  author={Shi, Modi and Peng, Shijia and Chen, Jin and Jiang, Haoran and Li, Yinghui and Huang, Di and Luo, Ping and Li, Hongyang and Chen, Li},
  journal={arXiv preprint arXiv:2602.10106},
  year={2026}
}

@article{he2025viral,
  title={VIRAL: Visual Sim-to-Real at Scale for Humanoid Loco-Manipulation},
  author={He, Tairan and Wang, Zi and Xue, Haoru and Ben, Qingwei and Luo, Zhengyi and Xiao, Wenli and Yuan, Ye and Da, Xingye and Casta{\~n}eda, Fernando and Sastry, Shankar and others},
  journal={arXiv preprint arXiv:2511.15200},
  year={2025}
}

@inproceedings{yu2025skillmimic,
  title={Skillmimic-v2: Learning robust and generalizable interaction skills from sparse and noisy demonstrations},
  author={Yu, Runyi and Wang, Yinhuai and Zhao, Qihan and Tsui, Hok Wai and Wang, Jingbo and Tan, Ping and Chen, Qifeng},
  booktitle={Proceedings of the Special Interest Group on Computer Graphics and Interactive Techniques Conference Conference Papers},
  pages={1--11},
  year={2025}
}

@inproceedings{wang2025skillmimic,
  title={Skillmimic: Learning basketball interaction skills from demonstrations},
  author={Wang, Yinhuai and Zhao, Qihan and Yu, Runyi and Tsui, Hok Wai and Zeng, Ailing and Lin, Jing and Luo, Zhengyi and Yu, Jiwen and Li, Xiu and Chen, Qifeng and others},
  booktitle={Proceedings of the IEEE/CVF Conference on Computer Vision and Pattern Recognition},
  pages={17540--17549},
  year={2025}
}

@article{li2026haic,
  title={HAIC: Humanoid Agile Object Interaction Control via Dynamics-Aware World Model},
  author={Li, Dongting and Chen, Xingyu and Wu, Qianyang and Chen, Bo and Wu, Sikai and Wu, Hanyu and Zhang, Guoyao and Li, Liang and Zhou, Mingliang and Xiang, Diyun and others},
  journal={arXiv preprint arXiv:2602.11758},
  year={2026}
}

@article{wu2026sugar,
  title={SUGAR: A Scalable Human-Video-Driven Generalizable Humanoid Loco-Manipulation Learning Framework},
  author={Wu, Tianshu and Kong, Xiangqi and Chen, Yue and Yu, Qize and Ye, Hang and Li, Jia and Wang, Yizhou and Dong, Hao},
  journal={arXiv preprint arXiv:2605.20373},
  year={2026}
}

@article{liu2025opt2skill,
  title={Opt2skill: Imitating dynamically-feasible whole-body trajectories for versatile humanoid loco-manipulation},
  author={Liu, Fukang and Gu, Zhaoyuan and Cai, Yilin and Zhou, Ziyi and Jung, Hyunyoung and Jang, Jaehwi and Zhao, Shijie and Ha, Sehoon and Chen, Yue and Xu, Danfei and others},
  journal={IEEE Robotics and Automation Letters},
  year={2025},
  publisher={IEEE}
}

@article{dong2026learning,
  title={Learning Humanoid End-Effector Control for Open-Vocabulary Visual Loco-Manipulation},
  author={Dong, Runpei and Li, Ziyan and He, Xialin and Gupta, Saurabh},
  journal={arXiv preprint arXiv:2602.16705},
  year={2026}
}

@article{he2024omnih2o,
  title={Omnih2o: Universal and dexterous human-to-humanoid whole-body teleoperation and learning},
  author={He, Tairan and Luo, Zhengyi and He, Xialin and Xiao, Wenli and Zhang, Chong and Zhang, Weinan and Kitani, Kris and Liu, Changliu and Shi, Guanya},
  journal={arXiv preprint arXiv:2406.08858},
  year={2024}
}

@article{sun2025ulc,
  title={Ulc: A unified and fine-grained controller for humanoid loco-manipulation},
  author={Sun, Wandong and Feng, Luying and Cao, Baoshi and Liu, Yang and Jin, Yaochu and Xie, Zongwu},
  journal={arXiv preprint arXiv:2507.06905},
  year={2025}
}

@article{xue2025opening,
  title={Opening the Sim-to-Real Door for Humanoid Pixel-to-Action Policy Transfer},
  author={Xue, Haoru and He, Tairan and Wang, Zi and Ben, Qingwei and Xiao, Wenli and Luo, Zhengyi and Da, Xingye and Casta{\~n}eda, Fernando and Shi, Guanya and Sastry, Shankar and others},
  journal={arXiv preprint arXiv:2512.01061},
  year={2025}
}

@article{ren2025humanoid,
  title={Humanoid Goalkeeper: Learning from Position Conditioned Task-Motion Constraints},
  author={Ren, Junli and Long, Junfeng and Huang, Tao and Wang, Huayi and Wang, Zirui and Jia, Feiyu and Zhang, Wentao and Wang, Jingbo and Luo, Ping and Pang, Jiangmiao},
  journal={arXiv preprint arXiv:2510.18002},
  year={2025}
}

@article{su2025hitter,
  title={Hitter: A humanoid table tennis robot via hierarchical planning and learning},
  author={Su, Zhi and Zhang, Bike and Rahmanian, Nima and Gao, Yuman and Liao, Qiayuan and Regan, Caitlin and Sreenath, Koushil and Sastry, S Shankar},
  journal={arXiv preprint arXiv:2508.21043},
  year={2025}
}

@article{zhang2026learning,
  title={Learning athletic humanoid tennis skills from imperfect human motion data},
  author={Zhang, Zhikai and Lu, Haofei and Lian, Yunrui and Chen, Ziqing and Liu, Yun and Lin, Chenghuai and Xue, Han and Zeng, Zicheng and Qi, Zekun and Zheng, Shaolin and others},
  journal={arXiv preprint arXiv:2603.12686},
  year={2026}
}

@article{mittal2025isaaclab,
  title={Isaac Lab: A GPU-Accelerated Simulation Framework for Multi-Modal Robot Learning},
  author={Mittal, Mayank and Roth, Pascal and Tigue, James and Richard, Antoine and Zhang, Octi and Du, Peter and Serrano-Mu{\~n}oz, Antonio and Yao, Xinjie and Zurbr{\"u}gg, Ren{\'e} and Rudin, Nikita and Wawrzyniak, Lukasz and Rakhsha, Milad and Denzler, Alain and Heiden, Eric and Borovicka, Ales and Ahmed, Ossama and Akinola, Iretiayo and Anwar, Abrar and Carlson, Mark T. and Feng, Ji Yuan and Garg, Animesh and Gasoto, Renato and Gulich, Lionel and Guo, Yijie and Gussert, M. and Hansen, Alex and Kulkarni, Mihir and Li, Chenran and Liu, Wei and Makoviychuk, Viktor and Malczyk, Grzegorz and Mazhar, Hammad and Moghani, Masoud and Murali, Adithyavairavan and Noseworthy, Michael and Poddubny, Alexander and Ratliff, Nathan and Rehberg, Welf and Schwarke, Clemens and Singh, Ritvik and Smith, James Latham and Tang, Bingjie and Thaker, Ruchik and Trepte, Matthew and Van Wyk, Karl and Yu, Fangzhou and Millane, Alex and Ramasamy, Vikram and Steiner, Remo and Subramanian, Sangeeta and Volk, Clemens and Chen, CY and Jawale, Neel and Kuruttukulam, Ashwin Varghese and Lin, Michael A. and Mandlekar, Ajay and Patzwaldt, Karsten and Welsh, John and Zhao, Huihua and Anes, Fatima and Lafleche, Jean-Francois and Mo{\"e}nne-Loccoz, Nicolas and Park, Soowan and Stepinski, Rob and Van Gelder, Dirk and Amevor, Chris and Carius, Jan and Chang, Jumyung and Chen, Anka He and de Heras Ciechomski, Pablo and Daviet, Gilles and Mohajerani, Mohammad and von Muralt, Julia and Reutskyy, Viktor and Sauter, Michael and Schirm, Simon and Shi, Eric L. and Terdiman, Pierre and Vilella, Kenny and Widmer, Tobias and Yeoman, Gordon and Chen, Tiffany and Grizan, Sergey and Li, Cathy and Li, Lotus and Smith, Connor and Wiltz, Rafael and Alexis, Kostas and Chang, Yan and Chu, David and Fan, Linxi "Jim" and Farshidian, Farbod and Handa, Ankur and Huang, Spencer and Hutter, Marco and Narang, Yashraj and Pouya, Soha and Sheng, Shiwei and Zhu, Yuke and Macklin, Miles and Moravanszky, Adam and Reist, Philipp and Guo, Yunrong and Hoeller, David and State, Gavriel},
  journal={arXiv preprint arXiv:2511.04831},
  year={2025},
  url={https://arxiv.org/abs/2511.04831}
}

@inproceedings{todorov2012mujoco,
  title={MuJoCo: A physics engine for model-based control},
  author={Todorov, Emanuel and Erez, Tom and Tassa, Yuval},
  booktitle={2012 IEEE/RSJ International Conference on Intelligent Robots and Systems},
  pages={5026--5033},
  year={2012},
  organization={IEEE},
  doi={10.1109/IROS.2012.6386109}
}

@article{wang2023physhoi,
  title={PhySHOI: Physics-Based Imitation of Dynamic Human-Object Interaction},
  author={Wang, Yinhuai and Lin, Jing and Zeng, Ailing and Luo, Zhengyi and Zhang, Jian and Zhang, Lei},
  journal={arXiv preprint arXiv:2312.04393},
  year={2023}
}

@article{lau2026switch,
  title={Switch: Learning Agile Skills Switching for Humanoid Robots},
  author={Lau, Yuen-Fui and Zhao, Qihan and Wang, Yinhuai and Yu, Runyi and Tsui, Hok Wai and Chen, Qifeng and Tan, Ping},
  journal={arXiv preprint arXiv:2604.14834},
  year={2026}
}

\clearpage
\appendix
\section*{Appendix}

\section{Dataset}
\label{sec:dataset}
We introduce the \textbf{OmniContact dataset}, a comprehensive human-object interaction (HOI) corpus tailored specifically for humanoid loco-manipulation. It captures object-constrained whole-body motions to supervise downstream policy learning. Unlike datasets using post-hoc labels, OmniContact directly pairs synchronized human motion with 6-DoF object trajectories, making the object state an intrinsic part of the record.

Capturing real physical interactions is critical, as behaviors like carrying, pushing, and kicking are heavily governed by object geometry and physical dynamics. The OmniContact dataset therefore emphasizes physically grounded, object-constrained motion clips that can be converted into contact-flow supervision for humanoid controllers.

\begin{figure*}[b]
    \centering
    \includegraphics[width=\linewidth]{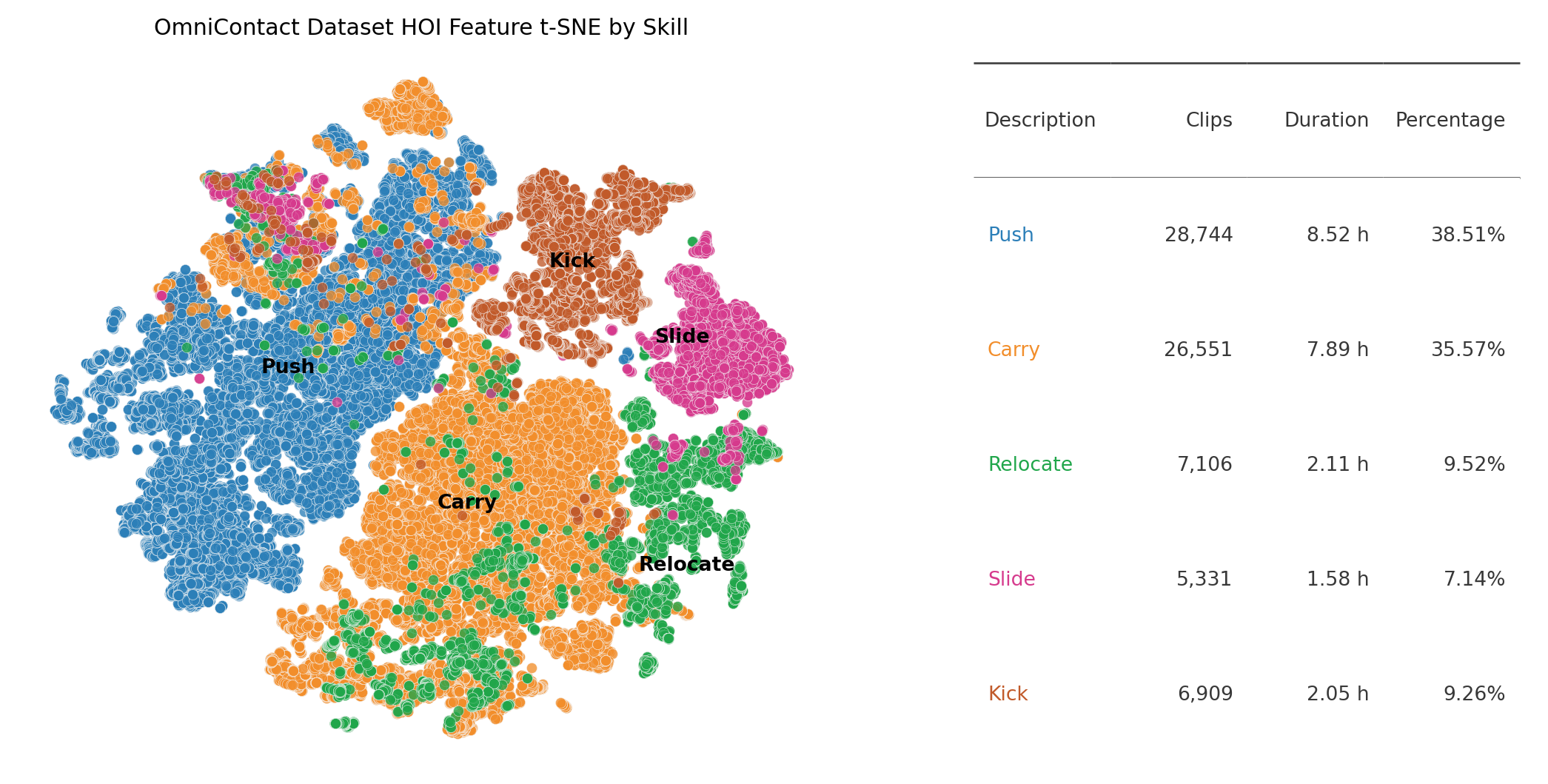}
    \caption{\textbf{Skill coverage of the OmniContact dataset.} We visualize HOI motion clips with t-SNE features and color them by primitive skill. The dataset covers both dominant long-horizon interactions, including \textit{Push} and \textit{Carry}, and specialized behaviors, including \textit{Relocate}, \textit{Slide}, and \textit{Kick}.}
    \label{fig:dataset_skill_tsne}
\end{figure*}

\begin{table*}[h]
    \centering
    \small
    \caption{\textbf{Dataset statistics comparison with OMOMO.} Metrics marked with $^\ast$ are computed on a 400-sequence OMOMO subset for per-trajectory statistics.}
    \label{tab:dataset_motion_statistics}
    \setlength{\tabcolsep}{4.5pt}
    \renewcommand{\arraystretch}{1.12}
    \resizebox{0.85\linewidth}{!}{
    \begin{tabular}{lcc}
        \toprule
        \textbf{Metric} & \textbf{OmniContact dataset} & \textbf{OMOMO} \\
        \midrule
        \multicolumn{3}{l}{\textbf{\textit{Dataset scale}}} \\
        Valid sequences & 1,274 & \textbf{6,435} \\
        Total motion duration & \textbf{22.29 h} & $\sim$10 h \\
        Total object frames & \textbf{7.22M} & $\sim$1.08M \\
        \midrule
        \multicolumn{3}{l}{\textbf{\textit{Representation and synchronization}}} \\
        Human motion representation & BVH motion & SMPL-X \\
        Object state representation & \textbf{Rigid-body 6-DoF} & Object pose \\
        Actor-object synchronization & \textbf{90 Hz paired capture} & 30 Hz paired sequence \\
        \midrule
        \multicolumn{3}{l}{\textbf{\textit{Loco-manipulation structure}}} \\
        Mean sequence duration & \textbf{62.98 s} & 5.69 s$^\ast$ \\
        Action primitives & carry / push / relocate / slide / kick & object manipulation \\
        Mean object path length & \textbf{19.76 m} & 2.67 m$^\ast$ \\
        Mean human root travel & \textbf{22.46 m} & 1.90 m$^\ast$ \\
        \midrule
        \multicolumn{3}{l}{\textbf{\textit{Interaction grounding}}} \\
        Contact timing resolution & \textbf{11.1 ms} & 33.3 ms \\
        Contact mode granularity & primitive-level contact mode & sequence-level interaction \\
        Dynamic/static ratio & 0.519 / 0.481 & 0.733 / 0.267$^\ast$ \\
        Object categories & 4 categories & \textbf{15 categories} \\
        Timestamp consistency & \textbf{100\%} & \textbf{100\%}$^\ast$ \\
        \bottomrule
    \end{tabular}}
\end{table*}

\paragraph{Interaction grounding.} 
OmniContact achieves precise actor-object synchronization by temporally aligning human kinematics with object trajectories. Instead of explicit force labels, we formulate contact observability through learning-friendly representations: paired trajectories, 6-DoF states, primitive-level contact modes, high-resolution timing, and dynamic/static phases. These features explicitly guide contact-flow learning on when, where, and how to interact with objects.

\paragraph{Comparison with OMOMO.}
While OMOMO offers greater sequence count and object diversity for short-window motion modeling, OmniContact focuses on longer, high-frequency demonstrations with extensive object transport. Specifically, OMOMO clips average 5.69 s and 2.67 m of travel, whereas OmniContact averages 62.98 s and 19.76 m. These datasets are complementary: OMOMO broadens object coverage, while OmniContact targets long-horizon loco-manipulation and contact-flow supervision for humanoid policy learning.

\paragraph{Coverage and annotation.} OmniContact is structured around reusable loco-manipulation primitives: carrying, pushing, kicking, relocating, and sliding. As visualized in Fig.~\ref{fig:dataset_skill_tsne}, HOI features form distinct clusters for specific skills alongside broad, overlapping regions. This distribution reveals both skill-specific patterns and shared motion modes, supporting our unified CF-Track policy. Furthermore, sequences are paired with verified task-level language descriptions detailing the primitive, object state, and target outcome.

\section{Task}
\label{sec:append_task}

This appendix details the task protocols for our loco-manipulation benchmark. We designed this evaluation suite to demonstrate that a streamlined binary contact abstraction (contact flow) can effectively represent a diverse set of everyday behaviors, including carrying, pushing, sliding, kicking, and stacking. To systematically assess our framework, the benchmark is hierarchically organized into two levels:
(1) \textit{Individual meta-skills}, which validate the acquisition of reusable contact modes; and
(2) \textit{Meta-skill chaining}, which challenges the agent to robustly sequence these skills over long horizons while adapting to dynamic object states and changing contact topologies.
In the following, we decompose each task into distinct phases driven by contact state transitions.

\subsection{Meta-Skill Tasks}
\label{sec:phase_templates}

\begin{itemize}
    \item \textbf{Carry Box.}
    \begin{itemize}
        \item \textit{Phase 1: Approach (No Contact).} The robot navigates toward the box to reach a feasible manipulation distance.
        \item \textit{Phase 2: Crouch and Grasp (No Contact).} The robot lowers its body and establishes stable hand contact with the box.
        \item \textit{Phase 3: Lift (In Contact).} The robot raises the box from the ground or table to a transport-ready height.
        \item \textit{Phase 4: Transport (In Contact).} The robot locomotes to the target location while maintaining continuous contact and whole-body balance.
        \item \textit{Phase 5: Crouch and Place (In Contact).} The robot lowers the box to the target location and transfers support back to the environment.
        \item \textit{Phase 6: Recover Standing (No Contact).} The robot returns to a nominal standing posture.
    \end{itemize}

    \item \textbf{Relocate Ball.}
    \begin{itemize}
        \item This task follows the same contact-flow structure as \textit{Carry Box} but targets a ground-initialized ball. 
    \end{itemize}

    \item \textbf{Push Suitcase.}
    \begin{itemize}
        \item \textit{Phase 1: Approach Waypoint (No Contact).} Navigates to an intermediate collision-free waypoint if the direct path to the object is obstructed.
        \item \textit{Phase 2: Approach Object (No Contact).} Moves to a pre-contact stance $0.4\,\mathrm{m}$ behind the suitcase, aligning the body with the intended pushing direction.
        \item \textit{Phase 3: Crouch and Contact (In Contact).} Establishes hand contact and transitions into a kinematically feasible pushing posture.
        \item \textit{Phase 4: Transport (In Contact).} Pushes the suitcase along a straight-line trajectory to the target destination.
        \item \textit{Phase 5: Recover Standing (No Contact).} Returns to a nominal standing posture.
    \end{itemize}

    \item \textbf{Slide Box.}
    \begin{itemize}
        \item \textit{Phase 1: Approach Waypoint (No Contact).} Navigates to an intermediate collision-free waypoint if the direct path to the object is obstructed.
        \item \textit{Phase 2: Approach Object (No Contact).} Moves to a pre-contact stance $0.2\,\mathrm{m}$ behind the suitcase, aligning the body with the intended sliding direction.
        \item \textit{Phase 3: Transport (In Contact).} The robot applies directional force to slide the box across the ground, actively controlling the object's motion until the target pose is reached.
    \end{itemize}

    \item \textbf{Kick Ball.}
    \begin{itemize}
         \item \textit{Phase 1: Approach Waypoint (No Contact).} Navigates to an intermediate collision-free waypoint if the direct path to the object is obstructed.
        \item \textit{Phase 2: Approach Object (No Contact).} Moves to a pre-contact stance $0.2\,\mathrm{m}$ behind the suitcase, aligning the body with the intended sliding direction.
        \item \textit{Phase 3: Strike (In Contact).} The robot executes a swift kicking motion, creating a brief but high-velocity contact with the ball.
        \item \textit{Phase 4: Recovery Standing (No Contact).} Returns to a nominal standing posture.
    \end{itemize}
\end{itemize}

\subsection{Meta-Skill Chaining Tasks}
Chaining tasks evaluate the robot's ability to sequence the aforementioned meta-skills, requiring deliberate breaking and re-establishing of contacts between distinct actions.

\begin{itemize}
    \item \textbf{Push-Stack Boxes.} The robot first executes \textit{Push Suitcase} to move a large box to a target destination, then transitions to \textit{Carry Box} to pick up and stack a smaller box on top of it.

    \item \textbf{Carry-Push Boxes.} The robot uses \textit{Carry Box} to place a small box onto a shelf, then switches to \textit{Push Suitcase} to maneuver a large box into the space beneath it.

    \item \textbf{Relocate-Kick Ball.} The robot executes \textit{Relocate Ball} to transport a ball to a penalty mark, recovers standing to break contact, and then repositions to execute \textit{Kick Ball} to score.

    \item \textbf{Push Box-Relocate Ball.} The robot uses \textit{Push Suitcase} to position a box in a target area, then transitions to \textit{Relocate Ball} to pick up a scattered ball and drop it inside the box.
\end{itemize}

\section{Additional Experiments}
\label{sec:appendix_exp}

\subsection{Locomotion Evaluation}
\label{sec:append_locomotion_eval}
As shown in Table~\ref{tab:locomotion_benchmark}, we compare the pure locomotion performance of OmniContact against several baselines across four diverse motion types. \textbf{OmniContact consistently outperforms all other methods}, achieving the lowest final torso position error (\(E^T_{\text{torso}}\)) of 0.199 and maintaining a perfect success rate (\(R_{\text{succ}}\)) of 100.0\% across all 83 tested motion files. In contrast, baselines struggle with specific motion types. Sonic demonstrates robust success rates but yields higher tracking errors compared to our method. These results highlight the superior stability and tracking accuracy of OmniContact in diverse locomotion scenarios.

\begin{table*}[h]
    \centering
    \small
    \caption{\textbf{Locomotion performance comparison.} Motion file counts are in parentheses.}
    \vspace{3pt}
    \label{tab:locomotion_benchmark}
    \setlength{\tabcolsep}{6pt}
    \renewcommand{\arraystretch}{1.2}
    \resizebox{\textwidth}{!}{
    \begin{tabular}{l cc cc cc cc | cc}
        \toprule
        \multirow{2}{*}{Method} & \multicolumn{2}{c}{Forward (48)} & \multicolumn{2}{c}{Backward (16)} & \multicolumn{2}{c}{Circle (16)} & \multicolumn{2}{c|}{Sideways (3)} & \multicolumn{2}{c}{Overall (83)} \\
        \cmidrule(lr){2-3} \cmidrule(lr){4-5} \cmidrule(lr){6-7} \cmidrule(lr){8-9} \cmidrule(lr){10-11}
        & \(E_{\text{torso}}^{T}\downarrow\) & \(R_{\text{succ}}\uparrow\) & \(E_{\text{torso}}^{T}\downarrow\) & \(R_{\text{succ}}\uparrow\) & \(E_{\text{torso}}^{T}\downarrow\) & \(R_{\text{succ}}\uparrow\) & \(E_{\text{torso}}^{T}\downarrow\) & \(R_{\text{succ}}\uparrow\) & \(E_{\text{torso}}^{T}\downarrow\) & \(R_{\text{succ}}\uparrow\) \\
        \midrule
        Sonic~\cite{luo2025sonic} & \(\underline{0.247}\) & \(\textbf{100.0}\) & \(\underline{0.239}\) & \(\textbf{100.0}\) & \(0.258\) & \(\textbf{100.0}\) & \(\underline{0.263}\) & \(\textbf{100.0}\) & \(\underline{0.248}\) & \(\textbf{100.0}\) \\
        HDMI~\cite{weng2025hdmi} & \(4.812\) & \(0.0\) & \(6.437\) & \(0.0\) & \(5.286\) & \(0.0\) & \(3.924\) & \(0.0\) & \(5.126\) & \(0.0\) \\
        PhysHSI~\cite{wang2025physhsi} & \(0.331\) & \(\textbf{100.0}\) & \(1.355\) & \(6.2\) & \(0.628\) & \(\underline{62.5}\) & \(0.356\) & \(\textbf{100.0}\) & \(0.586\) & \(81.9\) \\
        LessMimic~\cite{lin2026lessmimic} & \(0.300\) & \(\textbf{100.0}\) & \(14.332\) & \(\underline{75.0}\) & \(\underline{0.241}\) & \(\textbf{100.0}\) & \(0.266\) & \(\textbf{100.0}\) & \(2.992\) & \(\underline{96.4}\) \\
        \textbf{OmniContact} & \(\textbf{0.196}\) & \(\textbf{100.0}\) & \(\textbf{0.197}\) & \(\textbf{100.0}\) & \(\textbf{0.209}\) & \(\textbf{100.0}\) & \(\textbf{0.196}\) & \(\textbf{100.0}\) & \(\textbf{0.199}\) & \(\textbf{100.0}\) \\
        \bottomrule
    \end{tabular}}
\end{table*}

\subsection{Robustness Evaluation}
Table~\ref{tab:disturbance_replanning} evaluates the robustness of OmniContact against execution-time perturbations. We introduce two types of disturbances: (1) \textbf{Drop}, which simulates a severe failure by forcibly setting the box to the ground mid-carry during the \textit{Carry Box} task to explicitly trigger replanning; and (2) \textbf{Object Pose Offset}, which injects positional (\(\pm 10\)~cm in \(x, y\)) and rotational (\(\pm 90^\circ\)) noise into the object's pose immediately after the initial trajectory is planned. In all cases, CF-Gen effectively replans from the updated state and restores execution. The recovery is highly efficient, requiring only 1.5--1.8 replans on average while maintaining high success rates and low final object errors. These results demonstrate that our closed-loop pipeline provides a practical mechanism for recovering from unexpected physical disturbances.

\begin{table*}[h]
    \centering
    \small
    \caption{\textbf{Closed-loop replanning under disturbances.}}
    \vspace{3pt}
    \label{tab:disturbance_replanning}
    \setlength{\tabcolsep}{6pt}
    \renewcommand{\arraystretch}{1.12}
    \begin{tabular}{l | l | ccc}
        \toprule
        Task & Perturbation & Final Success (\%) & Avg. Replans & \(E_{\text{obj}}\downarrow\) \\
        \midrule
        Carry Box & Drop & 92.5 & 1.64 & 0.107 \\
        Carry Box & Object Pose Offset & 97.5 & 1.52 & 0.123 \\
        Push Suitcase & Object Pose Offset & 89.5 & 1.78 & 0.122 \\
        \bottomrule
    \end{tabular}
\end{table*}

\subsection{Extended Task Evaluation}
To demonstrate the versatility of OmniContact beyond the main benchmark, Table~\ref{tab:additional_omnicontact_tasks} evaluates additional meta-skills and a composed multi-stage task (e.g., sliding, kicking, and sequential relocation-plus-kick). These tasks encompass diverse object geometries and contact modes. We report open-loop success (\(R_{\text{succ}}\)), replanning-enabled success (\(R^*_{\text{succ}}\)), and final object error (\(E^T_{\text{obj}}\)). We observe lower success rates for ball-oriented tasks because the simulated ball is a smooth rigid sphere that easily slips during contact. We omit replanning success rates (\(R^*_{\text{succ}}\)) for \textit{Kick Ball} and its chaining tasks because the fast, low-friction ball rarely stops, making replanning unfeasible. Ultimately, these results highlight the flexibility of our contact-flow formulation across a broader range of humanoid loco-manipulation behaviors.

\begin{table*}[h]
    \centering
    \small
    \caption{\textbf{Additional OmniContact task performance.}}
    \vspace{3pt}
    \label{tab:additional_omnicontact_tasks}
    \setlength{\tabcolsep}{6pt}
    \renewcommand{\arraystretch}{1.12}
    \resizebox{0.7\textwidth}{!}{
    \begin{tabular}{llccc}
        \toprule
        Type & Task & \(R_{\text{succ}}(\%)\) & \(R_{\text{succ}}^{*}(\%)\) & \(E_{\text{obj}}^{T}\) \\
        \midrule
        \multirow{3}{*}{Meta-skill} & Relocate Ball & 72.5 & 89.0 & 0.35 \\
        & Slide Box & 81.5 & 92.4 & 0.24 \\
        & Kick Ball & 76.5 & - & 1.39 \\
        \midrule
        \multirow{2}{*}{Skill chaining} & Relocate-Kick Ball & 53.1 & - & 1.87 \\
        & Push Suitcase-Relocate Ball & 68.1 & 85.5 & 0.53 \\
        \bottomrule
    \end{tabular}
    }
\end{table*}

\subsection{Extended Long-Horizon Execution}
\label{sec:append_long_horizon}
As detailed in Table~\ref{tab:ultra_long_horizon_survival}, we expand our long-horizon evaluation across various protocols to test the system's stability. Each protocol introduces a different level of task complexity:

\begin{itemize}
    \item \textbf{Protocol I: Single-Object Sequential Goals.} This protocol tasks the agent with executing repeated sequential goals on a single object. Under this setting, OmniContact demonstrates exceptional stability, as all agents survive for 40 minutes with near-perfect replanning success and minimal object error.
    
    \item \textbf{Protocol II: Single-Object Task Resampling.} This protocol continuously modifies both the initial and final goals of the object. This introduces an additional navigation phase, requiring the robot to first approach the object before manipulation. Despite the increased difficulty of resetting states without environment resets, surviving agents maintain high success rates and low errors over long periods.
    
    \item \textbf{Protocol III: Multi-Object Sequential Goals.} In this most demanding scenario, the agent sequentially manipulates 5 varying-sized objects to their goals. The core challenge is managing inter-object interference (e.g., autonomously replanning to restore an already placed object that was accidentally displaced). This highlights the system's robustness to varying object sizes and environment-aware replanning capabilities.
\end{itemize}

\begin{table*}[h]
    \centering
    \small
    \caption{\textbf{Extended long horizon survival evaluation.}}
    \vspace{3pt}
    \label{tab:ultra_long_horizon_survival}
    \resizebox{\textwidth}{!}{%
    \setlength{\tabcolsep}{3.8pt}
    \renewcommand{\arraystretch}{1.15}
    \begin{tabular}{llccccc}
        \toprule
        & & & \multicolumn{4}{c}{Survival Performance} \\
        \cmidrule(lr){4-7}
        Protocol & Duration & Survival \(R_{\text{succ}}\) (\%) & \(E_{\text{obj}}^T\) & Avg. Skill Rounds & Avg. Object Drops & Avg. Replan \\
        \midrule
        \multirow{4}{*}{I} & 10 min & 100.0$^{\textcolor{gray}{(\pm 0.0)}}$ & 0.031$^{\textcolor{gray}{(\pm 0.0)}}$ & 271$^{\textcolor{gray}{(\pm 1.5)}}$ & 0.0$^{\textcolor{gray}{(\pm 0.0)}}$ & 0.0$^{\textcolor{gray}{(\pm 0.0)}}$ \\
         & 20 min & 100.0$^{\textcolor{gray}{(\pm 0.0)}}$ & 0.033$^{\textcolor{gray}{(\pm 0.0)}}$ & 551$^{\textcolor{gray}{(\pm 2.0)}}$ & 0.0$^{\textcolor{gray}{(\pm 0.0)}}$ & 0.0$^{\textcolor{gray}{(\pm 0.0)}}$ \\
         & 30 min & 100.0$^{\textcolor{gray}{(\pm 0.0)}}$ & 0.078$^{\textcolor{gray}{(\pm 0.1)}}$ & 805$^{\textcolor{gray}{(\pm 5.6)}}$ & 2.2$^{\textcolor{gray}{(\pm 5.4)}}$ & 88.6$^{\textcolor{gray}{(\pm 5.5)}}$ \\
         & 40 min & 100.0$^{\textcolor{gray}{(\pm 0.0)}}$ & 0.109$^{\textcolor{gray}{(\pm 0.2)}}$ & 1063$^{\textcolor{gray}{(\pm 12.0)}}$ & 3.5$^{\textcolor{gray}{(\pm 7.4)}}$ & 191.3$^{\textcolor{gray}{(\pm 7.5)}}$ \\
        \midrule
        \multirow{4}{*}{II} & 10 min & 73.5$^{\textcolor{gray}{(\pm 1.9)}}$ & 0.045$^{\textcolor{gray}{(\pm 0.0)}}$ & 136$^{\textcolor{gray}{(\pm 1.3)}}$ & 0.6$^{\textcolor{gray}{(\pm 1.5)}}$ & 5.4$^{\textcolor{gray}{(\pm 2.6)}}$ \\
         & 20 min & 38.5$^{\textcolor{gray}{(\pm 1.4)}}$ & 0.046$^{\textcolor{gray}{(\pm 0.0)}}$ & 121$^{\textcolor{gray}{(\pm 1.5)}}$ & 1.7$^{\textcolor{gray}{(\pm 2.9)}}$ & 4.8$^{\textcolor{gray}{(\pm 2.1)}}$ \\
         & 30 min & 31.0$^{\textcolor{gray}{(\pm 0.9)}}$ & 0.046$^{\textcolor{gray}{(\pm 0.0)}}$ & 184$^{\textcolor{gray}{(\pm 2.1)}}$ & 2.7$^{\textcolor{gray}{(\pm 3.8)}}$ & 5.5$^{\textcolor{gray}{(\pm 1.4)}}$ \\
         & 40 min & 29.5$^{\textcolor{gray}{(\pm 0.7)}}$ & 0.046$^{\textcolor{gray}{(\pm 0.0)}}$ & 247$^{\textcolor{gray}{(\pm 1.5)}}$ & 2.7$^{\textcolor{gray}{(\pm 3.8)}}$ & 4.9$^{\textcolor{gray}{(\pm 0.4)}}$ \\
        \midrule
        \multirow{4}{*}{III} & 10 min & 42.0$^{\textcolor{gray}{(\pm 11.2)}}$ & 1.103$^{\textcolor{gray}{(\pm 0.1)}}$ & 71$^{\textcolor{gray}{(\pm 1.5)}}$ & 4.7$^{\textcolor{gray}{(\pm 4.5)}}$ & 32.0$^{\textcolor{gray}{(\pm 2.4)}}$ \\
         & 20 min & 28.5$^{\textcolor{gray}{(\pm 12.0)}}$ & 0.985$^{\textcolor{gray}{(\pm 0.3)}}$ & 99$^{\textcolor{gray}{(\pm 6.4)}}$ & 9.0$^{\textcolor{gray}{(\pm 12.7)}}$ & 31.7$^{\textcolor{gray}{(\pm 1.6)}}$ \\
         & 30 min & 13.0$^{\textcolor{gray}{(\pm 0.0)}}$ & 1.063$^{\textcolor{gray}{(\pm 0.0)}}$ & 158$^{\textcolor{gray}{(\pm 0.0)}}$ & 25.0$^{\textcolor{gray}{(\pm 0.0)}}$ & 34.8$^{\textcolor{gray}{(\pm 0.8)}}$ \\
         & 40 min & 10.0$^{\textcolor{gray}{(\pm 0.0)}}$ & 1.034$^{\textcolor{gray}{(\pm 0.0)}}$ & 207$^{\textcolor{gray}{(\pm 0.0)}}$ & 26.0$^{\textcolor{gray}{(\pm 0.0)}}$ & 41.4$^{\textcolor{gray}{(\pm 0.3)}}$ \\
        \bottomrule
    \end{tabular}%
    }
\end{table*}

\subsection{Visualization Results}
\label{sec:append_visualization_results}
\begin{figure*}[h]
    \centering
    \subfigure[\textit{Carry Box}]{
        \includegraphics[width=0.98\linewidth]{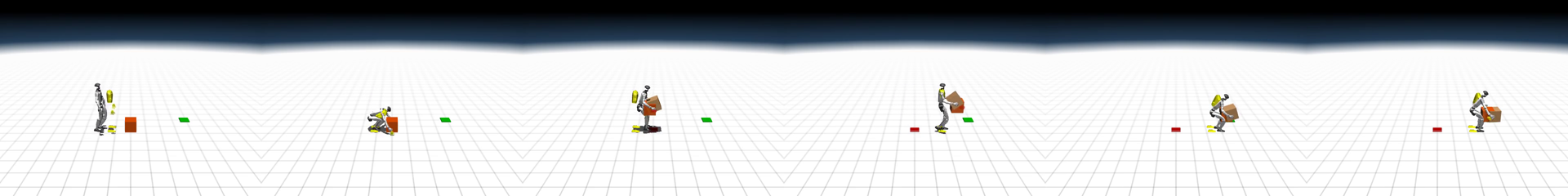}
    }
    \vspace{2pt}
    \subfigure[\textit{Push Suitcase}]{
        \includegraphics[width=0.98\linewidth]{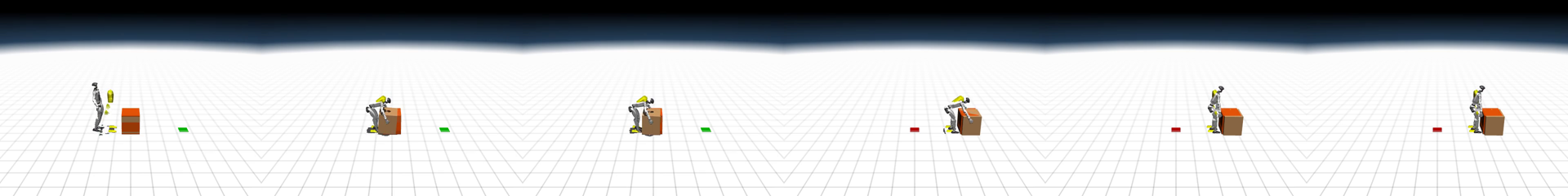}
    }
    \vspace{2pt}
    \subfigure[\textit{Relocate Ball}]{
        \includegraphics[width=0.98\linewidth]{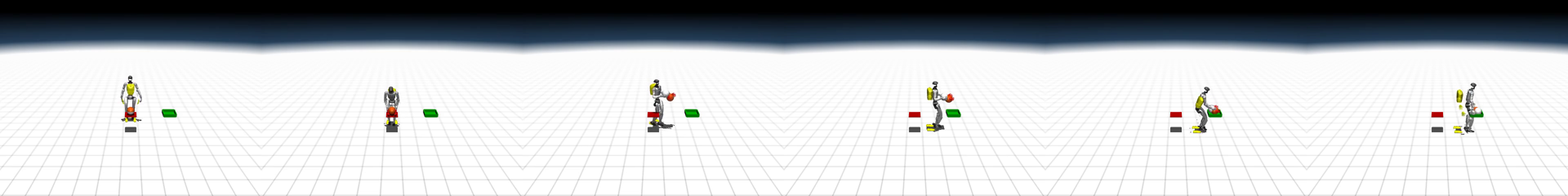}
    }
    \caption{\textbf{Qualitative rollouts of representative loco-manipulation skills.} Each strip contains six snapshots sampled from a successful execution, showing the temporal progression of humanoid-object interaction under \ours.}
    \label{fig:append_process_rollouts}
\end{figure*}

Fig.~\ref{fig:append_process_rollouts} visualizes representative successful rollouts for three loco-manipulation skills: \textit{Carry Box}, \textit{Push Suitcase}, and  \textit{Relocate Ball} . Each row contains six uniformly sampled snapshots from one execution, showing the transition from approaching the object, establishing contact, manipulating the object, and reaching the target state. These qualitative results complement the quantitative evaluations above by illustrating that \ours can maintain stable whole-body motion while adapting its contact pattern to different object geometries and task goals.

\section{Training Details}

\subsection{Experimental Settings}

All training experiments are conducted in Isaac Lab~\citep{mittal2025isaaclab} using four NVIDIA GeForce RTX 4090 GPUs. Unless otherwise specified, we simulate 4,096 parallel environments per GPU (16,384 in total), with a typical run converging in approximately 36 hours. Prior to real-world deployment, the learned policy is evaluated via sim-to-sim transfer in MuJoCo~\citep{todorov2012mujoco}.

For real-world demonstrations, we deploy the learned policy on the Unitree G1 humanoid robot. We use a Noitom motion-capture system and attach trackers to the robot pelvis and the manipulated object to obtain their global poses during deployment. The policy runs at 50 Hz, while the motion-capture system runs at 100 Hz.

\subsection{Training Configuration}
\label{sec:training_config}
\newcommand{\blankentry}{\strut}
\newcommand{\wideblankentry}{\strut}
\newcommand{\graybarc}[1]{\colorbox{black!12}{\makebox[\dimexpr\linewidth-2\fboxsep\relax][c]{\textbf{#1}}}}
\newcommand{\graybarl}[1]{\colorbox{black!12}{\makebox[\dimexpr\linewidth-2\fboxsep\relax][l]{\textbf{#1}}}}

In this section, we summarize the training configuration used for both the motion prior and the task-conditioned controller. Table~\ref{tab:training_hyperparameters} lists the optimization and network hyperparameters for AMPPPO training, while the AMP observation design and Table~\ref{tab:reward_terms} together specify the motion-imitation objective. For the policy network, the actor and critic are both Transformer-based rather than plain MLPs. For CF-Track, Table~\ref{tab:domain_randomization_tracking} reports the domain randomization ranges adopted for robust sim-to-real transfer, and Table~\ref{tab:observation_terms} summarizes the observation decomposition into tracking target, proprioceptive state, object state, and critic-only privileged state.

\begin{table*}[h]
    \centering
    \small
    \caption{Policy training hyperparameters.}
    \vspace{10pt}
    \label{tab:training_hyperparameters}
    \begin{minipage}[t]{0.31\linewidth}
        \centering
        \setlength{\tabcolsep}{4pt}
        \renewcommand{\arraystretch}{1.3}
        \resizebox{\linewidth}{!}{
        \begin{tabular}{lc}
            \toprule
            \multicolumn{2}{c}{\cellcolor{black!12}\textbf{General}} \\
            \textbf{Hyperparameter} & \textbf{Value} \\
            \midrule
            Algorithm & AMPPPO \\
            Runner & OnPolicyRunner2 \\
            Optimizer & Adam \\
            $\beta_1, \beta_2$ & $(0.9, 0.999)$ \\
            Learning Rate & $1.0 \times 10^{-3}$ \\
            LR Schedule & Adaptive \\
            Desired KL & $0.01$ \\
            Rollout Length & $24$ \\
            Rollout Batch Size & $98304$ \\
            Mini-batch Size & $24576$ \\
            Max Iterations & $50000$ \\
            Initial Action Noise Std & $1.0$ \\
            Observation Normalization & True \\
            \bottomrule
        \end{tabular}}
    \end{minipage}
    \hfill
    \begin{minipage}[t]{0.29\linewidth}
        \centering
        \setlength{\tabcolsep}{4pt}
        \renewcommand{\arraystretch}{1.0}
        \resizebox{\linewidth}{!}{
        \begin{tabular}{lc}
            \toprule
            \multicolumn{2}{c}{\cellcolor{black!12}\textbf{PPO Policy}} \\
            \textbf{Hyperparameter} & \textbf{Value} \\
            \midrule
            Discount Factor ($\gamma$) & $0.99$ \\
            GAE Parameter ($\lambda$) & $0.95$ \\
            Clip Parameter & $0.2$ \\
            Value Loss Coefficient & $1.0$ \\
            Clipped Value Loss & True \\
            Entropy Coefficient & $0.005$ \\
            Max Gradient Norm & $1.0$ \\
            Learning Epochs & $5$ \\
            Mini-batches & $4$ \\
            \midrule
            \multicolumn{2}{c}{\cellcolor{black!12}\textbf{Actor-Critic Transformer}} \\
            Hidden Dimensions & $[512, 256, 128]$ \\
            Layers &  $3$ \\
            Heads & $8$ \\ 
            Dropout & $0.0$ \\
            Activation & ELU \\
            \bottomrule
        \end{tabular}}
    \end{minipage}
    \hfill
    \begin{minipage}[t]{0.33\linewidth}
        \centering
        \setlength{\tabcolsep}{4pt}
        \renewcommand{\arraystretch}{1.3}
        \resizebox{\linewidth}{!}{
        \begin{tabular}{lc}
            \toprule
            \multicolumn{2}{c}{\cellcolor{black!12}\textbf{AMP}} \\
            \textbf{Hyperparameter} & \textbf{Value} \\
            \midrule
            AMP Discriminator MLP Size & $[256, 256]$ \\
            AMP Discriminator Optimizer & Adam \\
            AMP Discriminator LR & $1.0 \times 10^{-3}$ \\
            AMP Replay Buffer Size & $100000$ \\
            AMP Reward Coefficient & $0.5$ \\
            Task Reward Lerp & $0.85$ \\
            AMP Loss Coefficient & $1.0$ \\
            Gradient Penalty Coefficient & $1.0$ \\
            Gradient Penalty Lambda & $1.0$ \\
            AMP Observation Normalization & True \\
            AMP History Length & $10$ \\
            AMP Observation Dim & $410$ \\
            \bottomrule
        \end{tabular}}
    \end{minipage}
\end{table*}

\paragraph{AMP observation design.}
The AMP discriminator receives a 10-frame history, yielding an AMP observation dimension of 410. Each single-frame observation contains the following terms:
\begin{itemize}
    \item \texttt{base\_height} (1D), representing the base height.
    \item \texttt{joint\_pos} (29D), representing robot joint positions.
    \item \texttt{projected\_gravity} (3D), representing the gravity direction in the robot frame.
    \item \texttt{box\_pos\_local} (3D), representing the object position in the local robot frame.
    \item \texttt{contact\_info} (4D), representing binary contact indicators.
    \item \texttt{amp\_domain\_id} (1D), representing the AMP domain identifier.
    \item The resulting single-frame observation dimension is 41.
    \item We stack a history of 10 frames, resulting in a final AMP observation dimension of 410.
\end{itemize}

\begin{table*}[t]
    \centering
    \small
    \caption{Reward terms.}
    \label{tab:reward_terms}
    \setlength{\tabcolsep}{4pt}
    \renewcommand{\arraystretch}{1.15}
    \resizebox{\linewidth}{!}{
    \begin{tabular}{lccc}
        \toprule
        \textbf{Term} & \textbf{Expression} & \textbf{Weight} & \textbf{Remarks} \\
        \midrule
        \rowcolor{black!12}
        \multicolumn{4}{l}{\textbf{\textit{(a) Tracking Rewards}}} \\
        \midrule
        Torso position &
        $\exp(-\|\mathbf{p}^{ref}_{a}-\mathbf{p}_{a}\|^2 / 0.3^2)$ &
        $0.2$ & Torso anchor \\
        Torso orientation &
        $\exp(-d_q(\mathbf{q}^{ref}_{a}, \mathbf{q}_{a})^2 / 0.4^2)$ &
        $0.2$ & Torso anchor \\
        Local body position &
        $\exp(-\mathrm{mean}_{b\in\mathcal{B}}\|\mathbf{p}^{ref}_{b,rel}-\mathbf{p}_{b,rel}\|^2 / 0.3^2)$ &
        $0.2$ & All tracked bodies \\
        Local body orientation &
        $\exp(-\mathrm{mean}_{b\in\mathcal{B}} d_q(\mathbf{q}^{ref}_{b,rel}, \mathbf{q}_{b,rel})^2 / 0.4^2)$ &
        $0.2$ & All tracked bodies \\
        Wrist position &
        $\exp(-\mathrm{mean}_{b\in\mathcal{W}}\|\mathbf{p}^{ref}_{b}-\mathbf{p}_{b}\|^2 / 0.3^2)$ &
        $0.4$ & Left/right rubber hands \\
        Wrist orientation &
        $\exp(-\mathrm{mean}_{b\in\mathcal{W}} d_q(\mathbf{q}^{ref}_{b}, \mathbf{q}_{b})^2 / 0.4^2)$ &
        $0.4$ & Left/right rubber hands \\
        Object position &
        $\exp(-\|\mathbf{p}^{ref}_{o}-\mathbf{p}_{o}\|^2 / 0.3^2)$ &
        $1.0$ & Carried object \\
        Object orientation &
        $\exp(-d_q(\mathbf{q}^{ref}_{o}, \mathbf{q}_{o})^2 / 0.3^2)$ &
        $1.0$ & Carried object \\
        Interaction position &
        $\mathrm{mean}_{h\in\mathcal{H}}\exp(-\|\Delta\mathbf{p}_{h,o}-\Delta\mathbf{p}^{ref}_{h,o}\|^2 / \sigma_h^2)$ &
        $0.4$ & Hand-object relative position \\
        Interaction orientation &
        $\mathrm{mean}_{h\in\mathcal{H}}\exp(-(\Delta d_{h,o}-\Delta d^{ref}_{h,o})^2 / \sigma_h^2)$ &
        $0.4$ & Hand-object relative orientation \\
        Contact matching &
        $\exp(-\mathrm{MSE}(\mathbf{c}, \mathbf{c}^{ref}) / 1.0^2)$ &
        $1.0$ & Feet/hands contact labels \\
        \midrule
        \rowcolor{black!12}
        \multicolumn{4}{l}{\textbf{\textit{(b) Regularization and Safety Penalties}}} \\
        \midrule
        Action rate &
        $\|\mathbf{a}_{t}-\mathbf{a}_{t-1}\|^2$ &
        $-0.1$ & Smooth actions \\
        Elbow/wrist torque &
        $\sum_{j\in\mathcal{J}_{ew}}\tau_j^2$ &
        $-5.0\times10^{-3}$ & Elbow and wrist joints \\
        Joint limit &
        $\sum_j \max(q_j^{min}-q_j,0)+\max(q_j-q_j^{max},0)$ &
        $-10.0$ & Soft joint limits \\
        Foot slip &
        $\sum_{f\in\mathcal{F}}\mathbf{1}_{\mathrm{contact}}\|\mathbf{v}_{f,xy}\|$ &
        $-0.1$ & Ankles in contact \\
        \bottomrule
    \end{tabular}}
\end{table*}

\begin{table*}[t]
    \centering
    \begin{minipage}[t]{0.56\textwidth}
        \centering
        \small
        \caption{Domain randomization setting for CF-Track.}
        \label{tab:domain_randomization_tracking}
        \setlength{\tabcolsep}{6pt}
        \renewcommand{\arraystretch}{1.2}
        \resizebox{\linewidth}{!}{
        \begin{tabular}{ll}
            \toprule
            \textbf{Term} & \textbf{Value / Range} \\
            \midrule
            \multicolumn{2}{l}{\textbf{\textit{(a) External Disturbances}}} \\
            \midrule
            Push interval & \(1\)--\(3\) s \\
            Push velocity (\(v_x, v_y\)) & \(\mathcal{U}[-0.5, 0.5]\) m/s \\
            Push velocity (\(v_z\)) & \(\mathcal{U}[-0.2, 0.2]\) m/s \\
            \midrule
            \multicolumn{2}{l}{\textbf{\textit{(b) Robot Dynamics Randomization}}} \\
            \midrule
            Torso COM offset (\(x\)) & \(\mathcal{U}[-0.025, 0.025]\) m \\
            Torso COM offset (\(y, z\)) & \(\mathcal{U}[-0.05, 0.05]\) m \\
            Encoder bias & \(\mathcal{U}[-0.01, 0.01]\) \\
            Default joint position offset & \(\mathcal{U}[-0.01, 0.01]\) rad \\
            Joint stiffness and damping & log-uniform scale in \([0.75, 1.5]\) \\
            Rigid-body static friction & \(\mathcal{U}[0.3, 1.6]\) \\
            Rigid-body dynamic friction & \(\mathcal{U}[0.3, 1.2]\) \\
            Rigid-body restitution & \(\mathcal{U}[0.0, 0.5]\) \\
            \midrule
            \multicolumn{2}{l}{\textbf{\textit{(c) Object Dynamics Randomization}}} \\
            \midrule
            Static friction (push-task object) & \(\mathcal{U}[0.2, 0.4]\) \\
            Dynamic friction (push-task object) & \(\mathcal{U}[0.1, 0.3]\) \\
            Static friction (kick-task ball) & \(\mathcal{U}[0.06, 0.12]\) \\
            Dynamic friction (kick-task ball) & \(\mathcal{U}[0.04, 0.08]\) \\
            Static friction (relocation-task ball) & \(\mathcal{U}[0.5, 0.8]\) \\
            Dynamic friction (relocation-task ball) & \(\mathcal{U}[0.4, 0.7]\) \\
            Static friction (default objects) & \(\mathcal{U}[0.5, 0.8]\) \\
            Dynamic friction (default objects) & \(\mathcal{U}[0.3, 0.6]\) \\
            Object restitution & 0 \\
            Object mass & \(\mathcal{U}[0.5, 1.5] \times \text{default mass}\) \\
            Object inertia & recomputed after mass scaling \\
            \bottomrule
        \end{tabular}}
    \end{minipage}
    \hfill
    \begin{minipage}[t]{0.42\textwidth}
        \centering
        \small
        \caption{Observation terms for CF-Track.}
        \label{tab:observation_terms}
        \setlength{\tabcolsep}{4pt}
        \renewcommand{\arraystretch}{1.25}
        \resizebox{\linewidth}{!}{
        \begin{tabular}{lc}
            \toprule
            \textbf{State} & \textbf{Dim.} \\
            \midrule
            \rowcolor{black!12}\textbf{\textit{(a) Tracking Target}} & \\
            \midrule
            Reference Torso Position & $11 \times 3$ \\
            Reference Torso Orientation & $11 \times 6$ \\
            Reference Rubber Hand Position & $11 \times (2 \times 3)$ \\
            Reference Rubber Hand Orientation & $11 \times (2 \times 6)$ \\
            Reference Ankle Roll Position & $11 \times (2 \times 3)$ \\
            Reference Ankle Roll Orientation & $11 \times (2 \times 6)$ \\
            Reference Contact & $11 \times 4$ \\
            \midrule
            \rowcolor{black!12}\textbf{\textit{(b) Proprioceptive State}} & \\
            \midrule
            End-Effector Positions & $4 \times 3$ \\
            Base Angular Velocity & 3 \\
            Gravity Orientation & 3 \\
            Joint Position & 29 \\
            Joint Velocity & 29 \\
            Last Action & 29 \\
            \midrule
            \rowcolor{black!12}\textbf{\textit{(c) Object State}} & \\
            \midrule
            Object Position in Robot Frame & 3 \\
            Object Orientation in Robot Frame & 6 \\
            Object Bounding Box in Robot Frame& $8 \times 3$ \\
            \midrule
            \rowcolor{black!12}\textbf{\textit{(d) Critic-Only Privileged State}} & \\
            \midrule
            Torso Position Error & 3 \\
            Torso Orientation Error & 6 \\
            Base Linear Velocity & 3 \\
            \bottomrule
        \end{tabular}}
    \end{minipage}
\end{table*}

\section{Evaluation Details}
\subsection{Evaluation Protocol}
\label{sec:append_eval_protocol}

For each benchmark task, we evaluate all methods under randomized object initializations and target configurations. The sampled episodes are fixed across methods, so each method is tested on the same set of object poses, object geometries, and target states when the task is supported. Unless otherwise stated, each reported success rate is computed over 1000 evaluation episodes.

The baseline fairness subset is constructed from a pure MoCap-data subset with paired skill metadata. This subset fixes the same MoCap-derived initial object pose, waypoint sequence, and target object pose for each evaluated episode. For dense-tracking baselines such as Sonic~\cite{luo2025sonic} and HDMI~\cite{weng2025hdmi}, which require frame-level tracking commands, we convert the MoCap data and skill metadata into the dense tracking references required by their controllers. For goal-conditioned baselines such as PhysHSI~\cite{wang2025physhsi} and LessMimic~\cite{lin2026lessmimic}, we provide the supported sparse task inputs, including the MoCap-derived initial pose, intermediate waypoints, and target pose. For \ours, CF-Gen receives the same initial pose, waypoints, and target pose, and then synthesizes the contact-flow plan executed by CF-Track. Thus, the subset compares execution under matched MoCap-derived task metadata rather than evaluating baselines with weaker task information.

\begin{table*}[t]
    \centering
    \captionsetup{skip=3pt}
    \caption{\textbf{Baseline fairness subset in simulation}. Unlike the full randomized benchmark in Table~\ref{tab:simulation_benchmark}, this mean-only diagnostic subset is constructed from a pure MoCap-data subset and fixes matched initial poses, waypoints, and target poses for direct controller comparison. ``--'' denotes unsupported tasks.}
    \label{tab:baseline_fairness_subset}
    \setlength{\tabcolsep}{16pt}
    \renewcommand{\arraystretch}{1.02}
    \resizebox{0.7\textwidth}{!}{
        \begin{tabular}{l c c c c}
            \toprule
            \multirow{3}{*}{Methods} & \multicolumn{4}{c}{Meta-Skill Subset} \\
            \cmidrule(lr){2-5}
            & \multicolumn{2}{c}{\textit{Carry Box}} & \multicolumn{2}{c}{\textit{Push Suitcase}} \\
            \cmidrule(lr){2-3} \cmidrule(lr){4-5}
            & $R_{\text{succ}}(\%)\uparrow$ & $E_{\text{obj}}^{T}\downarrow$
            & $R_{\text{succ}}(\%)\uparrow$ & $E_{\text{obj}}^{T}\downarrow$ \\
            \midrule
            Sonic~\cite{luo2025sonic} & 
            $1.66$ & $2.18$
            & $0.00$ & $2.24$ \\
            HDMI~\cite{weng2025hdmi}
            & $0.00$ & $2.62$
            & $0.00$ & $2.43$ \\
            PhysHSI~\cite{wang2025physhsi}
            & $\underline{83.91}$ & $\underline{0.54}$
            & -- & -- \\
            LessMimic~\cite{lin2026lessmimic}
            & $38.00$ & $1.37$
            & $\underline{24.10}$ & $\underline{1.85}$ \\
            \textbf{OmniContact}
            & $\textbf{99.08}$ & $\textbf{0.05}$
            & $\textbf{86.30}$ & $\textbf{0.23}$ \\
            \bottomrule
        \end{tabular}
    }
\end{table*}


For the full benchmark, tracking-based baselines that require dense references, specifically Sonic and HDMI, are provided with the task-matched reference available under the same episode specification. When a MoCap-retargeted reference exists, we use it directly; otherwise, for tasks without direct MoCap demonstrations, we synthesize the required dense reference from the same task metadata used to define the episode. This setup is intentionally favorable to tracking baselines: it supplies them with episode-specific dense references rather than requiring them to plan contact-flow segments. The benchmark therefore evaluates whether each controller can faithfully execute the same randomized interaction episodes, while the long-horizon chaining columns in the main paper additionally reflect whether a method supports the required task interface.

\begin{itemize}
    \item \textbf{Carry Box.} We sample 1000 different initial box poses. The box center is sampled with $x,y \in [-5, 5]$, and the vertical position is sampled from the box half-height to $0.8$. To test shape diversity, the box dimensions are randomized independently, with length, width, and height each sampled from $[0.20, 0.50]$. A trial is successful if the robot establishes contact with the box, lifts or stably carries it, and moves it near the target location.

    \item \textbf{Push Suitcase.} We evaluate 1000 randomized suitcase-pushing episodes. The suitcase initial position and the final goal are randomized in the ground plane, and the robot must first align the suitcase with the goal direction by rotating around its root. After alignment, the suitcase must be pushed along the specified straight trajectory to the final goal. We randomize the suitcase pose and target direction to test both contact establishment and heading control.

    \item \textbf{Stack Boxes.} We evaluate 1000 stacking episodes with three boxes initialized at randomized ground-plane positions. The goal is fixed as a vertical stack at the final target region, while the initial box ordering and spatial layout vary across episodes. A trial is successful only when all three boxes are moved to the goal region and stacked along the $z$ axis in the required order.

    \item \textbf{Push-Stack Boxes.} We evaluate 1000 skill-chaining episodes that combine suitcase pushing and box stacking. The suitcase and box initial poses are randomized, and the robot must first push the suitcase to its destination before transitioning to box manipulation. The task succeeds only if the suitcase reaches the target and the small box is subsequently stacked on top of it.

    \item \textbf{Additional meta-skills.} For additional skills such as \textit{Slide Box}, \textit{Kick Box}, and \textit{Kick Ball}, we follow the same randomized evaluation principle: object poses and target states are sampled across episodes, and success requires achieving the intended object displacement while keeping the humanoid balanced. These tasks are used to test whether the binary contact-flow interface generalizes beyond carrying and pushing.
\end{itemize}

\subsection{Naturalness Score Evaluation}
\label{sec:append_naturalness_eval}

To evaluate the naturalness score \(N_{\text{hoi}}\), we utilize Gemini-3.1-Pro as a zero-shot vision-based evaluator, as we found its assessments to be highly consistent with human evaluation results. The model is prompted to assess the robot manipulation videos based solely on visual evidence, remaining strictly blind to file names, method identities, or any other contextual metadata. The evaluation criteria are tailored to the specific dynamics of each task:

\begin{itemize}
    \item \textbf{Carry Box}: Evaluates whether the robot establishes physical contact, maintains a stable grasp throughout the transport phase, and successfully navigates to the target location.
    \item \textbf{Push Suitcase}: Evaluates whether the robot makes natural contact with the suitcase, applies appropriate and continuous force to push it, and smoothly navigates toward the goal.
    \item \textbf{Slide Box}: Evaluates whether the robot establishes realistic foot contact, executing step-by-step kicks to maintain a continuous sliding motion toward the final goal.
    \item \textbf{Kick Ball}: Evaluates whether the robot executes a natural kicking motion, makes accurate foot contact with the ball, and successfully kicks it in the intended direction.
\end{itemize}

Notably, since the LessMimic baseline lacks a box-release action, the evaluator is explicitly instructed to overlook this omission, provided the robot stably carries the box to the destination. The prompt used for this evaluation is detailed below. The evaluator is configured to return one row of assessment per video, with the format illustrated in Table~\ref{tab:naturalness_prompt_output}.

\begin{lstlisting}[language={},escapeinside={(*@}{@*)}]
You are an evaluator for robot manipulation and simulation videos. Judge only from the video content. Do not use the file name, method name, or any prior knowledge.

Task: The robot should contact the box, lift/carry it, and move it near the target location. Note: LessMimic does not have a box-release behavior. For LessMimic, do not penalize the video for not putting the box down; it is sufficient if the robot stably carries the box near the target.

Please assign a Naturalness Score from 0 to 10. Higher means the motion is more natural, stable, and physically plausible.

Scoring guide:
- 9-10: Very natural motion. Stable walking, reasonable box contact, smooth box motion, almost no visible jitter, sliding, or coordination issues.
- 7-8: Overall natural and stable. Completes the carry with only minor jitter, posture issues, or slightly unnatural contact.
- 5-6: Acceptable but clearly imperfect. Noticeable stiffness, slight box sliding, unstable gait, or abrupt motion.
- 3-4: Poor motion but successful contact. Completes the carry, but accompanied by severe jitter, abnormal posture, extreme stiffness, or obvious body instability.
- 1-2: Unsuccessful carry but natural motion. Fails to establish plausible contact or lift the box, yet maintains stable walking, smooth reaching, and reasonable kinematics.
- 0: Complete failure. Fails to contact or lift the box, combined with highly unnatural body motion, severe instability, or falling. Also includes invalid videos.

Consider:
1. Walking stability: balance and gait quality.
2. Box contact: whether the hands/body contact the box in a plausible carrying pose.
3. Box stability: whether the box moves smoothly without obvious sliding, bouncing, penetration, or falling.
4. Motion smoothness: absence of sudden jitter, joint twitching, or velocity discontinuities.
5. Task-level naturalness: whether the robot moves the box near the target in a reasonable way.

Output a table in the following format:
Video ID | Success Valid | Naturalness Score | Main Reason
\end{lstlisting}

\begin{table*}[h]
    \centering
    \small
    \captionsetup{skip=2pt}
    \caption{Example output format for the naturalness-score evaluator on the \textit{Carry Box} task.}
    \label{tab:naturalness_prompt_output}
    \setlength{\tabcolsep}{5pt}
    \renewcommand{\arraystretch}{1.15}
    \begin{tabular}{l c c p{0.47\linewidth}}
        \toprule
        \textbf{Video ID} & \textbf{Success Valid} & \textbf{Naturalness Score} & \textbf{Main Reason} \\
        \midrule
        0.mp4 & yes & 6 & The robot successfully picks up and carries the box to the target, but the walking is somewhat stiff. \\
        1.mp4 & yes & 8 & Overall natural and stable motion. The walking, lifting, carrying, and placing of the box are smooth with only minor imperfections. \\
        2.mp4 & yes & 4 & The robot moves the box to the target by throwing it, but the throwing action is highly abrupt and unnatural, featuring sudden velocity changes and jerky torso/arm movements. \\
        3.mp4 & no & 0 & The robot fails to complete the task because it does not achieve a stable lift-and-carry of the box, making the video invalid for naturalness evaluation. \\
        \bottomrule
    \end{tabular}
\end{table*}

\section{Compatibility with VLMs}
\label{sec:vlm_integration}

The compact and structured representation of contact flow provides a natural interface for high-level semantic planners, such as vision-language models (VLMs). A VLM can decompose complex tasks into discrete object-level subgoals, which CF-Gen then grounds into executable contact-flow segments for CF-Track to execute.

This abstraction avoids requiring foundation models to predict dense humanoid kinematics directly. By restricting the VLM's reasoning to object-level planning and delegating low-level, contact-rich execution to \ours, the system can handle semantically grounded long-horizon tasks such as structured object rearrangement and multi-stage manipulation.

We show this capability via two task types:
\begin{itemize}
    \item[\textbf{(1)}] \textbf{Language-grounded transfer:} The VLM performs open-vocabulary and attribute-based visual reasoning. This enables tasks like identifying a cylinder, or selecting the soccer ball.
    \item[\textbf{(2)}] \textbf{Concept-driven layout:} The VLM translates abstract semantic concepts into precise geometric configurations, such as arranging scattered objects into a ``heart'' shape.
\end{itemize}

\begin{figure}[h]
    \centering
    \subfigure[Move the cylinder to the basket.]{
        \includegraphics[width=0.98\linewidth]{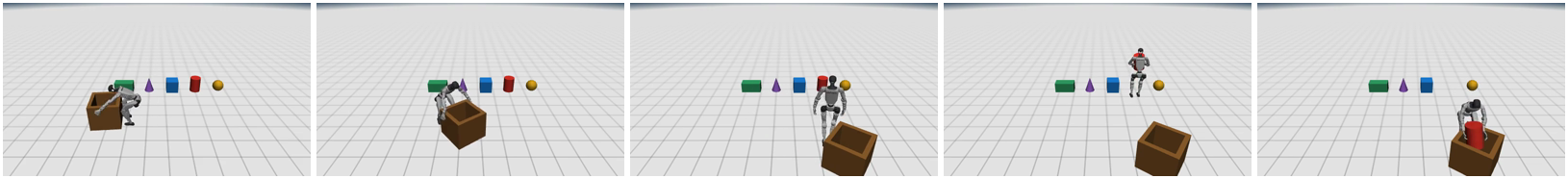}
    }
    \subfigure[Move the black-and-white soccer ball to the goal.]{
        \includegraphics[width=0.98\linewidth]{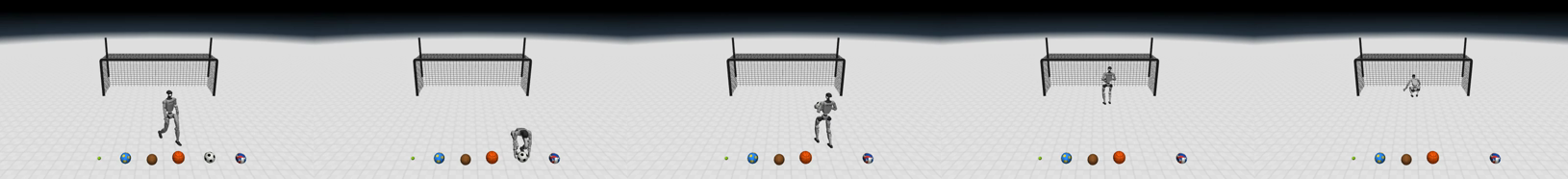}
    }
    \subfigure[Arrange objects as a heart.]{
        \includegraphics[width=0.98\linewidth]{fig/vlm_examples/heart_progress.png}
    }
    \subfigure[Arrange objects as ``Noitom'' with color-matched box targets.]{
        \includegraphics[width=0.98\linewidth]{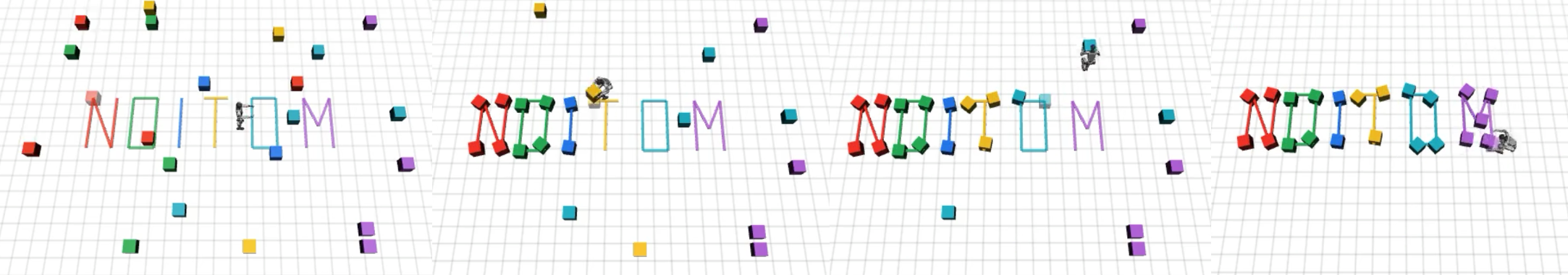}
    }
    \caption{\textbf{Progress visualizations from VLM-guided planning rollouts.}
    Each row shows five frames sampled from a demonstration video in temporal order. The first two examples require language-grounded object selection and goal assignment, while the last two require concept-driven spatial decomposition into object-level target poses. The ``Noitom'' task additionally requires matching each box to the target location with same color.}
    \label{fig:vlm_video_examples}
\end{figure}

Fig.~\ref{fig:vlm_video_examples} shows progress visualizations extracted from our VLM-guided rollouts, where each row contains temporally ordered frames from left to right. Given a scene observation and a natural-language instruction, the VLM first identifies task-relevant objects and converts the semantic request into an ordered list of object goals. For object-transfer tasks, this plan specifies which object should be moved and the target receptacle or region. For spatial-layout tasks, the VLM additionally generates a set of target poses that instantiate the requested concept, such as a heart contour or the letters ``Noitom''. These object-level goals are then passed to CF-Gen, which selects the corresponding meta-skill and synthesizes contact-flow segments for CF-Track. 

\paragraph{Qualitative protocol.}
We use Gemini 3.1 Pro Preview as the high-level VLM planner in our qualitative evaluation. The VLM receives a rendered top-down scene observation, the natural-language task instruction, and a concise description of the available meta-skills supported by CF-Gen. We constrain the VLM output to object-level planning only: it must identify the relevant object instances, assign each object a target region or target pose, and select one of the available meta-skills. The VLM is not asked to predict humanoid joint trajectories, contact timings, or low-level actions. These physical details are generated by CF-Gen and executed by CF-Track.

\begin{lstlisting}[language={},escapeinside={(*@}{@*)},caption={Prompt template for VLM-guided object-level planning.}]
You are a high-level planner for a humanoid loco-manipulation system.
Input:
1. A top-down image of the scene with movable objects.
2. A natural-language task instruction.
3. Available meta-skills: pick-place, push, kick, and spatial rearrangement.

Your job:
- Identify the task-relevant objects from the image.
- Convert the instruction into object-level subgoals.
- For each subgoal, choose a meta-skill and specify the target pose or target region.
- Do not output humanoid joint motions, contact timings, or low-level controls.

Return the plan using the required JSON schema.
\end{lstlisting}
\vspace{1.5em}

\begin{lstlisting}[language={},escapeinside={(*@}{@*)},caption={Required VLM output schema passed to CF-Gen.}]
{
  "task_type": "object_transfer | spatial_rearrangement",
  "subgoals": [
    {
      "object_id": "<visible object identifier>",
      "visual_attributes": {
        "shape": "<optional object shape>",
        "color": "<optional object color or texture>"
      },
      "skill": "pick_place | push | kick",
      "target": {
        "type": "region | pose",
        "position_xy": [x, y],
        "yaw": theta,
        "semantic_label": "<optional target description>",
        "matching_constraint": "<optional visual constraint, e.g., same color>"
      }
    }
  ]
}
\end{lstlisting}

\begin{table*}[h]
    \centering
    \small
    \caption{\textbf{Qualitative VLM planning tasks.}
    We evaluate whether the VLM can convert semantic instructions into object-level goals that can be consumed by CF-Gen.}
    \label{tab:vlm_task_protocol}
    \setlength{\tabcolsep}{5pt}
    \renewcommand{\arraystretch}{1.12}
    \begin{tabular}{p{0.26\linewidth} p{0.30\linewidth} p{0.34\linewidth}}
        \toprule
        Category & Example instruction & Expected object-level plan \\
        \midrule
        Language-grounded transfer & Move the cylinder to the basket. & Select the cylindrical object, assign the basket region as the target, and invoke a push or pick-place segment. \\
        Language-grounded transfer & Move the black-and-white patterned soccer ball to the goal. & Select the soccer ball by its visual texture, assign the goal region as the target, and invoke a kick or push segment. \\
        Concept-driven layout & Arrange the objects into a heart shape. & Generate a set of target poses distributed along a heart contour. \\
        Concept-driven layout & Arrange the objects into ``Noitom'' while matching each box to the target with the same color. & Generate target poses that form the requested letters and assign each box to a target location with the corresponding color. \\
        \bottomrule
    \end{tabular}
\end{table*}

\begin{figure*}[h]
    \centering
    \subfigure[Ambiguous object grounding.]{
        \includegraphics[width=0.31\linewidth]{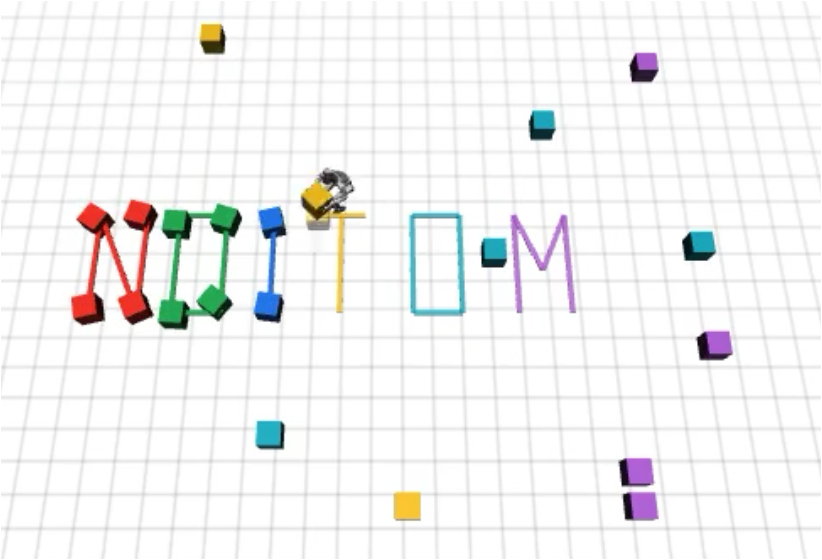}
    }
    \subfigure[Infeasible spatial layout.]{
        \includegraphics[width=0.31\linewidth]{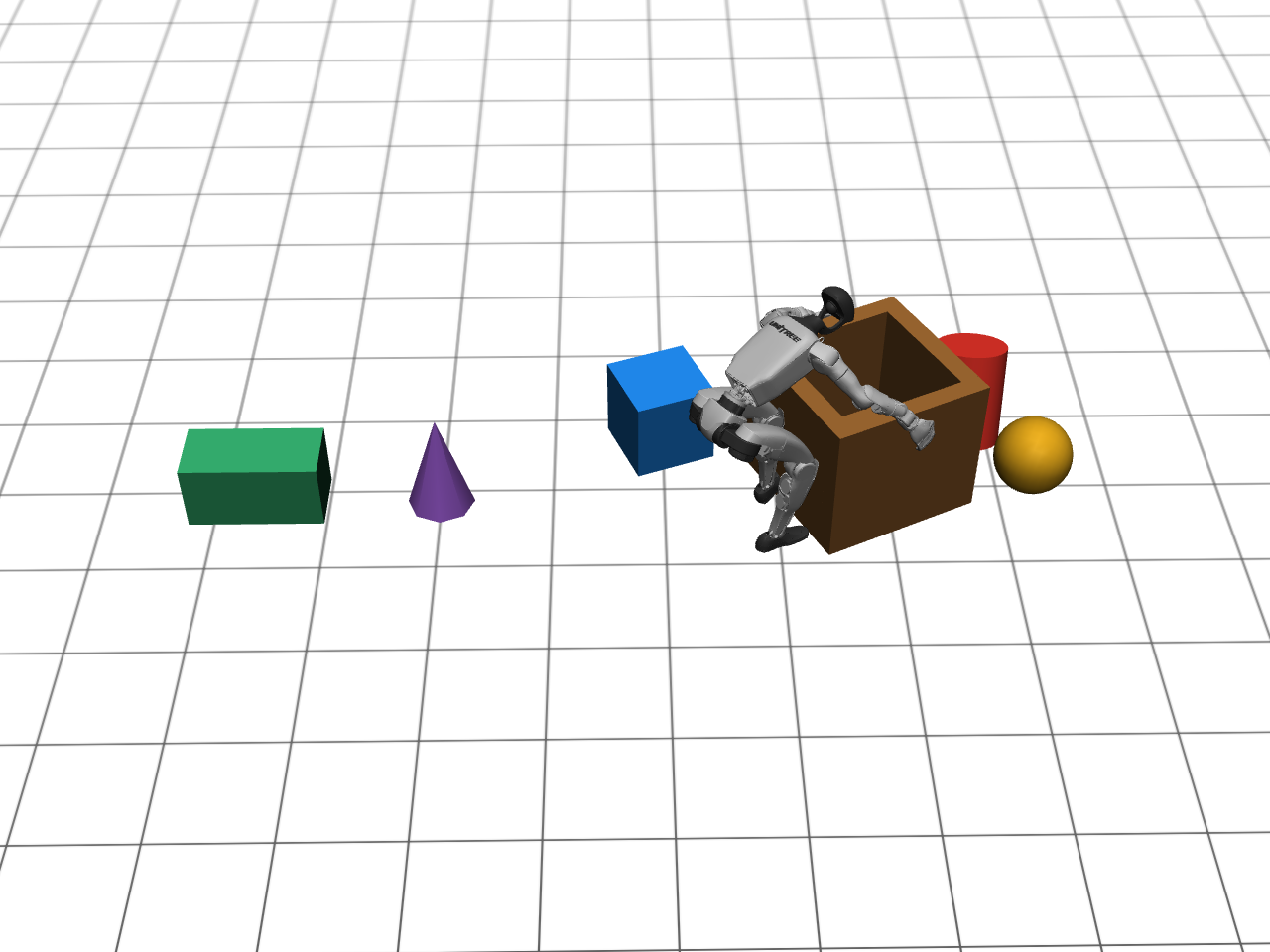}
    }
    \subfigure[Execution-induced deviation.]{
        \includegraphics[width=0.31\linewidth]{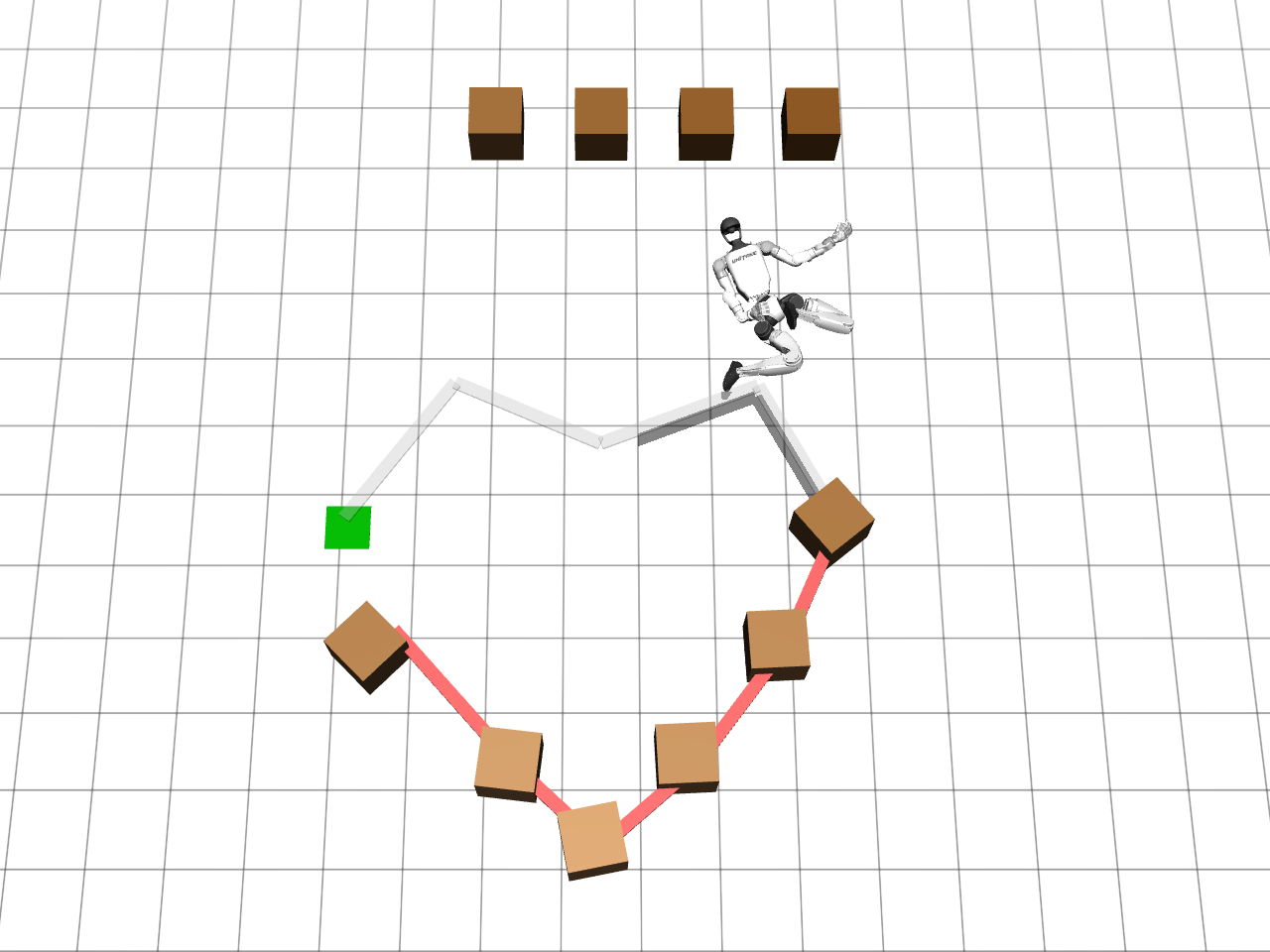}
    }
    
    \caption{\textbf{VLM-related failure cases.}
    Representative failures include placing a box intended for the ``R'' target onto the ``O'' target with mismatched colors, pushing the basket into the cylinder and displacing it instead of first moving the basket near the cylinder for pickup, and a low-level execution failure where excessive robot rotation causes a fall.}
    \label{fig:vlm_failure_cases}
\end{figure*}

\end{document}